\documentclass{article}

\usepackage[preprint]{neurips_2025}

\usepackage[utf8]{inputenc} % allow utf-8 input
\usepackage[T1]{fontenc}    % use 8-bit T1 fonts
\usepackage{hyperref}       % hyperlinks
\usepackage{url}            % simple URL typesetting
\usepackage{booktabs}       % professional-quality tables
\usepackage{amsfonts}       % blackboard math symbols
\usepackage{nicefrac}       % compact symbols for 1/2, etc.
\usepackage{microtype}      % microtypography

\usepackage{graphicx}
\usepackage{arydshln}
\usepackage{multirow}
\usepackage{caption}
\usepackage{subcaption}
\usepackage{makecell}
\usepackage{csquotes}
\usepackage{epigraph}
\usepackage{amsmath}
\usepackage{cleveref}
\usepackage{pifont}
\usepackage{listings}
\usepackage{amssymb}
\usepackage{wrapfig}

\usepackage{colortbl}
\usepackage[table,xcdraw]{xcolor}

\definecolor{lightpurple}{RGB}{230, 230, 250}

\newcommand\our{MoTE}
\newcommand{\cmark}{\ding{51}}%
\newcommand{\xmark}{\ding{55}}%

\definecolor{sgreen}{RGB}{30, 150, 30} 
\definecolor{green}{RGB}{0,210,0}
\definecolor{myblue}{RGB}{218,232,252}
\definecolor{mygray}{RGB}{220,220,220}
\definecolor{mypink}{RGB}{251,220,220}

\title{\our{}: Mixture of Ternary Experts for Memory-efficient Large Multimodal Models}

\author{%
  Hongyu Wang$^{\dagger \ddagger}$~~~Jiayu Xu$^{\dagger \ddagger}$~~~Ruiping Wang$^{\dagger \ddagger}$~~~Yan Feng$^\S$~~~Yitao Zhai$^\S$ \\
  \bf Peng Pei$^\S$~~~Xunliang Cai$^\S$~~~Xilin Chen$^{\dagger \ddagger}$ \\
  $^{\dagger}$Key Laboratory of AI Safety, Institute of Computing Technology, Chinese Academy of Sciences \\
  $^\S$Independent Researcher~~~$^{\ddagger}$University of Chinese Academy of Sciences
}

\begin{document}

\maketitle

%%%%%%%%% ABSTRACT
\begin{abstract}
Large multimodal Mixture-of-Experts (MoEs) effectively scale the model size to boost performance while maintaining fixed active parameters. However, previous works primarily utilized full-precision experts during sparse up-cycling. Despite they show superior performance on end tasks, the large amount of experts introduces higher memory footprint, which poses significant challenges for the deployment on edge devices. In this work, we propose \textbf{\our{}}, a scalable and memory-efficient approach to train \textbf{M}ixture-\textbf{o}f-\textbf{T}ernary-\textbf{E}xperts models from dense checkpoint. Instead of training fewer high-precision experts, we propose to train more low-precision experts during up-cycling. Specifically, we use the pre-trained FFN as a shared expert and train ternary routed experts with parameters in \{-1, 0, 1\}. Extensive experiments show that our approach has promising scaling trend along model size. \our{} achieves comparable performance to full-precision baseline MoE-LLaVA while offering lower memory footprint. Furthermore, our approach is compatible with post-training quantization methods and the advantage further amplifies when memory-constraint goes lower. Given the same amount of expert memory footprint of 3.4GB and combined with post-training quantization, \our{} outperforms MoE-LLaVA by a gain of 4.3\% average accuracy on end tasks, demonstrating its effectiveness and potential for memory-constrained devices.

\end{abstract}

%%%%%%%%% BODY TEXT
\section{Introduction}

Large Multimodal Models (LMMs)~\cite{phi3, mm1, mm1_5, qwen2vl, internvl2, qwen2_5_vl} have achieved remarkable performance across a wide range of downstream tasks, including visual question answering and autonomous computer agents. However, as model size increases, the rising inference cost presents significant challenges for deploying LMMs efficiently. To address this, Mixture-of-Experts (MoE)~\cite{gshard, switch, deepseekv3} introduces a mechanism that maintains a large pool of experts while activating only a subset for each input, thereby improving computational efficiency. Although MoE models significantly reduce FLOPs, they generally have a higher memory footprint, making deployment on edge devices challenging. For example, when training multimodal MoE up-cycled from Qwen2.5-3B, \textbf{if all feed-forward network (FFN) layers are replaced with MoE layers containing 16 experts, the resulting model's non-embedding memory footprint will increase from 5.2GB to 73.2GB.} This limitation is particularly pronounced for consumer-grade GPUs, which often have constrained memory capacities.

Model quantization is a promising approach to reducing the memory footprint of LMMs while maintaining comparable performance. Most mainstream quantization methods~\cite{gptq, awq, quip, qtip} aim to compress the bit-width of a pre-trained, full-precision model. Although these methods have a low training cost, they suffer from significant performance degradation when the bit-width is reduced below 4-bit. Recent studies~\cite{bitnetx, spectra, matmul_free} have demonstrated promising scaling trends for ternary pre-training in Large Language Models (LLMs). At sufficiently large model sizes, ternary models can achieve accuracy comparable to full-precision models on downstream tasks while maintaining the same pre-training cost. Furthermore, they have much lower inference costs in terms of memory, latency, and energy consumption~\cite{bitnetcpp}. However, since these models have only been trained on billions of tokens, a substantial performance gap remains between open-sourced ternary models and full-precision dense models. As a result, directly training MoE models initialized from these under-trained models leads to weak performance on end tasks.

In this work, we introduce \textbf{\our{}}, a scalable and memory-efficient architecture designed to train \textbf{M}ixture-\textbf{o}f-\textbf{T}ernary \textbf{E}xperts model from a pre-trained, full-precision dense checkpoint in multimodal tuning. Our approach addresses the inefficiency of multimodal MoE models in terms of memory footprint. Prior works~\cite{moe_llava, unimoe} primarily replace the FFN layer in dense checkpoints with an MoE layer, initializing the experts using the pre-trained FFN. However, we observed that in ternary training, replacing the FFN layer leads to significant performance degradation, as weight ternarization disrupts the pre-trained FFN. To mitigate this, we retain the FFN from the dense checkpoint as a shared expert activated for all inputs. During up-cycling, the layers inherited from the dense model remain frozen, while only the ternary MoE layers are trainable.

We first conduct strict and controlled experiments to evaluate the proposed approach against full-precision up-cycling MoE-LLaVA~\cite{moe_llava} across various model scales on a wide range of image understanding tasks. Our results show that ternary up-cycling exhibits surprising effectiveness as model size scales. 
As the size of the up-cycled dense checkpoint increases, the performance gap between \our{} and MoE-LLaVA narrows, eventually reaching comparable performance at scales larger than 1.5 billion parameters. 
Additionally, \our{} is compatible with post-training quantization techniques~\cite{gptq}. Given the same expert memory footprint and combined with post-training quantization, \our{} outperforms full-precision MoE-LLaVA at both 1.5B and 3B model sizes. This advantage becomes even more pronounced as memory constraints tighten. Specifically, under an expert memory budget of 3.4GB, our approach achieves a 4.3\% improvement in average accuracy on downstream task. 
These results demonstrate that given the same amount of total memory footprint and active parameter counts, training with a larger number of low-precision experts yields better performance than using fewer high-precision experts.

\section{Related Work}
\paragraph{Mixture of Experts.} LMMs demonstrate superior performance across various tasks as model size and training data scale increase. MoE models~\cite{gshard, switch, olmoe, deepseekvl2} maintain a large pool of experts but activate only a subset for each token, enabling improved performance at the same FLOPs budget. \cite{upcycling} introduced sparse up-cycling to reduce the training costs of MoE models by initializing them from dense checkpoints. \cite{moe_llava} explored the up-cycling of LMMs in the context of multimodal training, while \cite{llava_mod} proposed a progressive knowledge transfer strategy to train small-scale multimodal MoEs from dense models. \cite{unimoe} presented a scalable multimodal model that utilizes MoE with modality-specific encoders. While previous~\cite{moe_llava, cumo, unimoe} primarily focused on full-precision experts for up-cycling, our work investigates up-cycling with ternary experts to develop memory-efficient multimodal MoE models.

\paragraph{Model Quantization.} Quantization is a promising approach to reducing the memory footprint of LMMs while maintaining competitive performance, which can be categorized into two types based on the stage at which it is applied: post-training~\cite{llmint8, gptq, awq, quip, quip_sharp, qtip} and pre-training quantization~\cite{bitnet, bitnetx, fp4, fp8}. Post-training quantization compresses high-precision pre-trained models after training. Due to its lower cost, it is widely adopted for mainstream large-scale models. GPTQ~\citep{gptq} and AWQ~\citep{awq} reduce the bit-width to 4 bits while incurring minimal degradation. QuIP\#~\citep{quip_sharp} builds on QuIP~\citep{quip} by improving incoherence processing and applying vector quantization to incoherent weights. With additional fine-tuning, QuIP\# achieves state-of-the-art performance in 2-bit models. However, when the bit-width is reduced below 4-bit, these methods all suffer from significant performance degradation compared to BF16 baselines. In contrast, pre-training quantization integrates quantization into the training process, requiring models to be trained from scratch, which results in better performance. Recent~\cite{bitnetx} showed that ternary LLMs match the performance of full-precision counterpart starting from 3B parameter counts. \cite{qmoe} quantized a 1.6 trillion parameter Switch Transformer to sub 1-bit precision. \citep{moe-quant} proposed to quantize the experts with a mixed precision recipe and introduced a novel data-driven techniques for optimizing bit allocation.

\section{\our{}: Mixture-of-Ternary-Experts}
\label{sec:method}

In this section, we provide an overview of the proposed \our{}, including model architecture in Section~\ref{sec:arch}, training recipe in Section~\ref{sec:recipe} and objectives in Section~\ref{sec:obj}.

\subsection{Architecture}
\label{sec:arch}

\begin{figure*}[t]
    \centering
    \includegraphics[width=\textwidth]{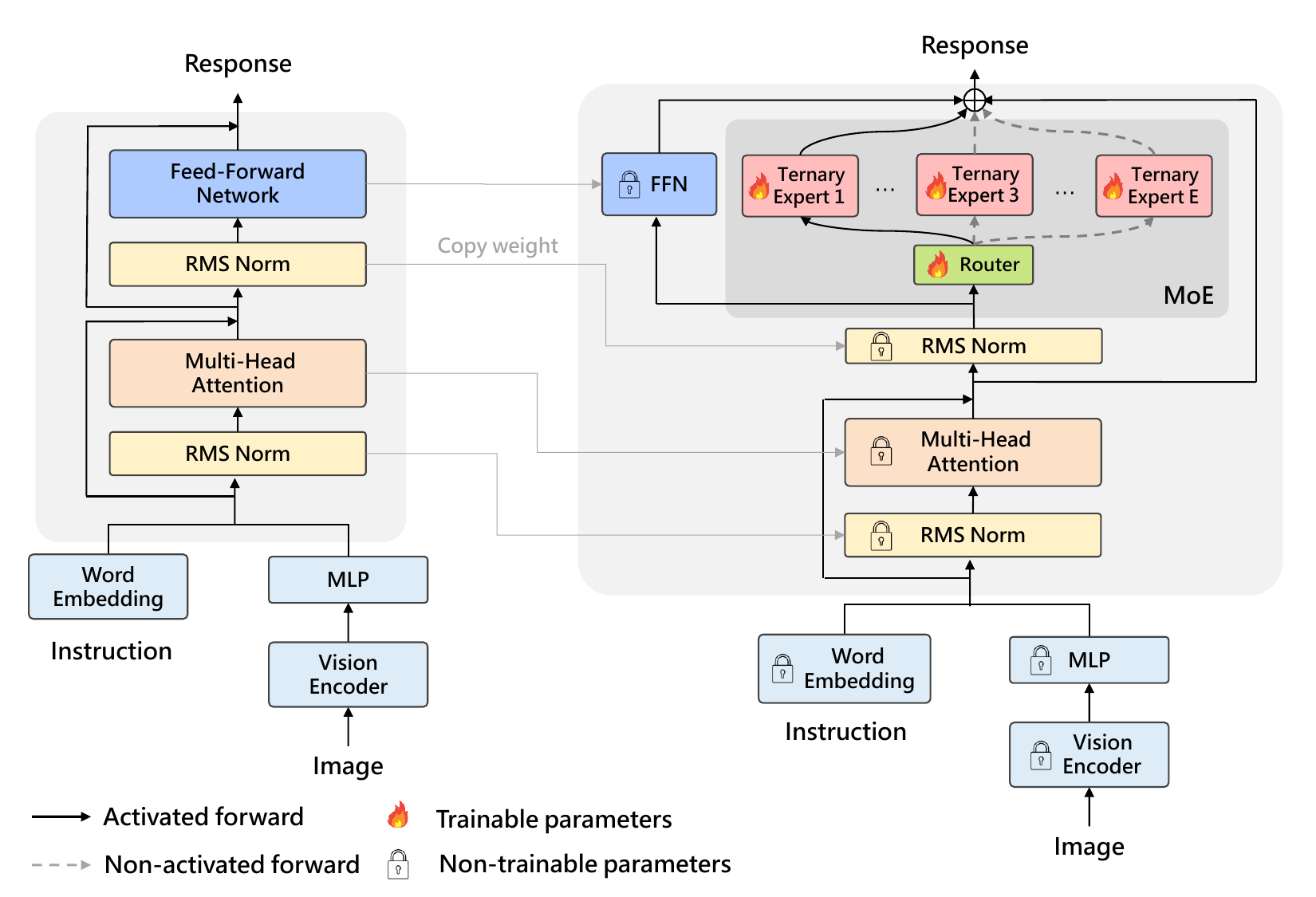}
    \caption{The overview of \our{}. We retain the pre-trained full-precision FFN as a shared expert and add a top-1 activated MoE layer with ternary experts. All experts and attention layers are initialized from the dense checkpoint.}
    \label{fig:overview}
\end{figure*}

We illustrate the architecture of \our{} in Figure~\ref{fig:overview}. Previous studies~\cite{upcycling, moe_llava} expanded a dense model into an MoE model by directly replacing the FFN layer with an MoE layer, where each expert is initialized from the dense FFN to accelerate convergence. However, as shown in Table~\ref{tab:bit_shard}, we found that directly replacing the FFN with an MoE in ternary up-cycling leads to significant performance degradation. We hypothesize that this occurs because the FFN encodes a substantial amount of factual knowledge acquired during pre-training~\cite{tran_are_kv, knowledge_neuron}, and weight ternarization severely disrupts pre-trained information. To mitigate this issue, we retain the FFN module from the dense model as a shared expert, ensuring it is activated for every token. Specifically, the forward computation of the $l$-th layer of \our{} can be formulated as:
\begin{align}
    &x_{l}^a = x_{l-1} + \text{MSA}(\text{LN}(x_{l-1})) \\
    &x_{l} = x_{l}^a + \text{MoE}(\text{LN}(x_{l}^a)) + \text{FFN}(\text{LN}(x_{l}^a))
\end{align}
where MSA and LN stands for multi-head self-attention and layer normalization, respectively. As illustrated in Figure~\ref{fig:overview}, we initialize the FFN, MSA and MoE layers from the dense model. We implement the MoE mechanism following the GShard~\cite{gshard}, with each expert modeled as a Gated Linear Unit (GLU)~\cite{glu}. An MoE layer which consists of $E$ ternary experts $\text{FFN}^T_1$ ... $\text{FFN}^T_E$ satisfies that:
\begin{align}
    &\mathcal{P}(x)_i = \cfrac{e^{f(x)_i}}{\sum_{j=1}^E e^{f(x)_j}} \label{eq:gate} \\
    &\text{MoE}(x) = \sum_{i=1}^E \mathcal{P}(x)_i \cdot \text{FFN}_i^T(x)
\end{align}
where $f(x)$ is the gating logits produced by the router. We leave the projection in router as BF16, since it only accounts for very small portion of total memory footprint. The forward computation of the $i$-th ternary expert $\text{FFN}_i^T(x)$ satisfies that:
\begin{align}
    &\text{FFN}_i^T(x) = Q_w (W_{\text{down}}^T) Q_a(h)\\
    &h = Q_w (W_{\text{up}}^T) Q_a(x) \otimes \sigma[Q_w(W_{\text{gate}}^T) Q_a(x)]
\end{align}
$\sigma$ is SiLU function. We apply \textit{absmean} quantizer and \textit{per-token absmax} quantizer for weight and activation quantization in expert's linear layers following BitNet~\cite{bitnetx}. Specifically, the quantization can be formulated as:
\begin{align}
    &Q_w(W) = \alpha \cdot \mathrm{RoundClip}(\frac{W}{\alpha}, -1, 1), \\
    &Q_a(x) = \frac{\beta}{127} \cdot \mathrm{RoundClip}(\frac{127x}{\beta}, -128, 127) \\
    &\alpha = \frac{1}{nm} || W||_1, \quad \beta = ||x||_{\infty} \\
    &\mathrm{RoundClip}(x, a, b) = \max(a, \min(b, \mathrm{round}(x)))
\end{align}
The weight $W \in \mathcal{R}^{m\times n}$ is quantized into ternary values, i.e., $\{-1, 0, 1\}$. The activations $x$ are per-token quantized into 8-bit integers, i.e., $[-128, 127]$. The output of ternary linear layer $Y$ is $Q_w(W) Q_a(x)$. During inference, we use the kernel from BitBlas~\cite{ladder} to save the memory footprint and accelerate the inference. Despite ternary values results in 1.58-bit, i.e., $\log 3 / \log 2$, BitBlas still stores and processes ternary weight in INT2 format since current GPUs are still based on binary system.

\subsection{Training recipe}
\label{sec:recipe}
Following MoE-LLaVA~\cite{moe_llava}, the training of \our{} consists of three stages. In Stage I, we train a two-layer MLP connector to align the visual encoder and LLM. As for Stage II, we fine-tune the LLM and connector using more complex vision-language instruction data. In Stage III, we expand the dense model from Stage II to an MoE model with ternary experts. The visual encoder is frozen through the training process. As presented in Figure~\ref{fig:overview}, during up-cycling, only ternary MoE layers are trainable, and the shared expert and MSA layers are frozen.

We adopt quantization-aware training for \our{}. The weights and activations are quantized into ternary and INT8 values on-the-fly. Since many operations in the quantization are no-differentiable, we deploy straight-through estimator~\cite{ste} for gradient approximation. The gradients are directly by-passing through non-differentiable functions, i.e., $\frac{\partial \mathcal{L}}{\partial W} = \frac{\partial \mathcal{L}}{\partial Q(W)}$ and $\frac{\partial \mathcal{L}}{\partial X} = \frac{\partial \mathcal{L}}{\partial Q(X)}$. The gradients and optimizer states are retained as full-precision.

\subsection{Training objectives}
\label{sec:obj}

The training objective of \our{} $\mathcal{L}_{\text{total}}$ requires the minimization of both the loss of specific multimodal tasks $\mathcal{L}_{\text{LM}}$ and an auxiliary load balancing loss $\mathcal{L}_{\text{balance}}$.

\paragraph{Language modeling loss.} The auto-regressive language modeling loss $\mathcal{L}_{\text{LM}}$ is widely adopted in the training of LMMs. Specifically, let $\mathcal{V}$ and $\mathcal{T}$ denote sequences of visual tokens and textual tokens, respectively. $\mathcal{T}$ can be divided as the instruction part $\mathcal{T}_{ins}$ and the response part $\mathcal{T}_{ans}$. The language modeling loss is calculated as:
\begin{equation}
    \mathcal{L}_{\text{LM}} = - \sum_{\text{token}_i \in \mathcal{T}_{ans}} \log \Pr (\mathcal{Y}^i \,|\,\mathcal{V}, \mathcal{T}^{[:i-1]})
\end{equation}
where $\mathcal{Y}$ is the model's output. We only calculate the loss on the response part.
\paragraph{Load balancing loss.} To ease the expert load imbalance problem in MoE, we adopt an auxiliary loss following Switch Transformers~\cite{switch}. Given a batch of training tokens $\mathbf{X}$, the balancing loss can be formulated as:
\begin{equation}
    \mathcal{L}_{\text{balance}} = \cfrac{E}{|\mathbf{X}|} \sum_{i=1}^E \sum_{x \in \mathbf{X}} t_i \cdot \mathcal{P}(x)_i
\end{equation}
where $|\mathbf{X}|$ is the number of training tokens in $\mathbf{X}$, $\mathcal{P}(x)_i$ is the routing logits depicted in Equation~\ref{eq:gate}, $t_i$ is the number of tokens routed to the $i$-th expert.

Above all, the training objective of \our{} is:
\begin{equation}
    \mathcal{L}_{\text{total}} = \mathcal{L}_{\text{LM}} + \gamma \cdot \mathcal{L}_{\text{balance}}
\end{equation}
where $\gamma$ is a coefficient for load balancing.

\section{Experiments}
\label{sec:exp}

\subsection{Setup}
\label{sec:setup}

\begin{wraptable}{r}{8cm}
    \centering
    \small
    \caption{The active/total parameter counts and expert memory of \our{} and MoE-LLaVA in various model sizes.}
    \label{tab:compare}
    \scalebox{0.85}{
    \begin{tabular}{l|ccc|c}
    \toprule
    \multirow{2}{*}{\bf Method}    &  \multicolumn{3}{c|}{\bf \# Active/Total Params} & \bf Expert \\ 
    & Stage I & Stage II & Stage III & \bf Memory $\downarrow$ \\
    \midrule
    \multicolumn{5}{l}{\ \emph{0.5B Model Up-cycling}} \\
    MoE-LLaVA & \multirow{2}{*}{1B} & \multirow{2}{*}{1B} & 1.3B/1.8B & 2.3GB (2.55$\times$) \\ 
    \our{} & & & 1.3B/2.1B & \bf 0.9GB (1.00$\times$) \\
    \midrule
    \multicolumn{5}{l}{\ \emph{1.5B Model Up-cycling}} \\
    MoE-LLaVA & \multirow{2}{*}{2B} & \multirow{2}{*}{2B} & 3.1B/5.4B & 8.6GB (2.69$\times$) \\ 
    \our{} & & & 3.1B/6.6B & \bf 3.2GB (1.00$\times$) \\
    \midrule
    \multicolumn{5}{l}{\ \emph{3B Model Up-cycling}} \\
    MoE-LLaVA & \multirow{2}{*}{3.4B} & \multirow{2}{*}{3.4B} & 5.9B/10.8B & 18.1GB (2.66$\times$) \\
    \our{} & & & 5.9B/13.2B & \bf 6.8GB (1.00$\times$) \\
    \bottomrule
    \end{tabular}
    }
\end{wraptable}

\paragraph{Model settings.} We select MoE-LLaVA~\cite{moe_llava} as the baseline. It adopts a similar three-stage MoE training recipe and utilizes full-precision experts. Since MoE-LLaVA activates the top-2 experts, and our model includes a shared expert, we use top-1 gating in \our{} to ensure a fair comparison in terms of FLOPs. All MoE layers consist of four routed experts. We adopt SigLIP-L~\cite{siglip} as the vision encoder and the instruct-version of Qwen2.5-series model~\cite{qwen2_5} as the base LLM. The connector is a two-layer MLP with GELU activation. Table~\ref{tab:compare} presents the active and total parameter counts in the training of \our{} and MoE-LLaVA across different model sizes. The expert memory footprint includes contributions from both shared and routed experts.

\paragraph{Implementation details.} We adopt expert parallelism for efficient training of MoE models. The coefficient $\gamma$ for load balancing loss is set as 0.01. The value is recommended by \cite{switch} to ensure auxiliary loss not to overwhelm the primary language modeling objective. Due to the limited computation resources, we do not perform dynamic resolution processing for the images, since it leads to extremely long training sequence. The length of the total sequence is set as 2048 tokens, and the visual input includes 729 tokens. More hyper-parameters can be found in Appendix~\ref{ap:hyper}.

\paragraph{Training data.} We train \our{} and MoE-LLaVA on the same dataset to ensure a fair comparison. The training dataset consists of a total of 5 million samples. For the first stage, we use the pre-training data of LLaVA 1.5~\cite{llava_1_5}. For the second stage, we use the mixture of SViT~\cite{svit}, LVIS~\cite{lvis}, LRV~\cite{lrv} and MIMIC-IT~\cite{mimic}. For the third stage, we use a subset of MAmmoTH-VL~\cite{mammothvl}, which includes 3.4 million instruction-response pairs, each associated with a single image as the visual input.

\paragraph{Evaluation.} We report the zero-shot performance of these models on a range of image understanding tasks using LMM-Eval toolkit~\cite{lmm-eval}, including MMMU~\cite{mmmu}, MathVista~\cite{mathvista} (MathV), MMBench~\cite{mmbench} (MMB), MMStar~\cite{mmstar} (MMS), MM-Vet~\cite{mmvet} (MMV), SeedBench-2-Plus~\cite{seedplus} (Seed$^{2+}$), SeedBench~\cite{seedbench} (Seed), AI2D~\cite{ai2d}, ChartQA~\cite{chartqa}, InfoVQA~\cite{infovqa} and DocVQA~\cite{docvqa}.

\begin{table*}[t]
    \small
    \caption{The results of \our{} and MoE-LLaVA on image understanding tasks in different model sizes. All models utilize the same base LLM, vision encoder and training dataset to ensure a fair comparison.}
    \label{tab:main}
    \begin{center}
        \scalebox{0.75}{
        \begin{tabular}{l|cccccccccc}
        \toprule
        \bf Method & \makecell{\bf MMMU \\ (val)} & \makecell{\bf MathV \\ (testmini)} & \makecell{\bf MMB \\ (en test)} & \makecell{\bf MMS \\ (test)} & \makecell{\bf Seed$^{2+}$ \\ (test)} & \makecell{\bf AI2D \\ (test)} & \makecell{\bf ChartQA \\ (test)} & \makecell{\bf InfoVQA \\ (val)} & \makecell{\bf DocVQA \\ (val)} & \bf Avg. \\
        \midrule
        \multicolumn{6}{l}{\ \emph{0.5B Model Up-cycling}} \\
        MoE-LLaVA & 35.4 & 35.4 & 57.3 & 39.5 & 43.3 & 57.4 & 56.0 & 25.8 & 49.3 & 44.4 \\
        \bf \our{} & 34.2 & 35.2 & 57.6 & 37.9 & 44.8 & 55.2 & 54.9 & 25.2 & 49.7 & 43.8 \\
        \bf \textcolor{black!60}{$\Delta$ \textit{compare to MoE-LLaVA}} & \textcolor{blue}{-1.2$\!\!$} & \textcolor{blue}{-0.2$\!\!$} & \textcolor{red}{+0.3$\!\!$} & \textcolor{blue}{-1.6$\!\!$} & \textcolor{red}{+1.5$\!\!$} & \textcolor{blue}{-2.2$\!\!$} & \textcolor{blue}{-1.1$\!\!$} & \textcolor{blue}{-0.6$\!\!$} & \textcolor{red}{+0.4$\!\!$} & \textcolor{blue}{-0.6$\!\!$} \\
        \midrule
        \multicolumn{6}{l}{\ \emph{1.5B Model Up-cycling}} \\
        MoE-LLaVA & 41.2 & 41.7 & 68.4 & 45.0 & 52.9 & 67.8 & 59.4 & 31.8 & 55.1 & 51.5 \\
        \bf \our{} & 42.6 & 44.8 & 70.0 & 46.4 & 54.8 & 68.7 & 61.3 & 32.5 & 57.4 & 53.2 \\
        \bf \textcolor{black!60}{$\Delta$ \textit{compare to MoE-LLaVA}} & \textcolor{red}{+1.4$\!\!$} & \textcolor{red}{+3.1$\!\!$} & \textcolor{red}{+1.6$\!\!$} & \textcolor{red}{+1.4$\!\!$} & \textcolor{red}{+1.9$\!\!$}& \textcolor{red}{+0.9$\!\!$} & \textcolor{red}{+1.9$\!\!$} & \textcolor{red}{+0.7$\!\!$} & \textcolor{red}{+2.3$\!\!$} & \textcolor{red}{+1.7$\!\!$} \\
        \midrule
        \multicolumn{6}{l}{\ \emph{3B Model Up-cycling}} \\
        MoE-LLaVA & 42.3 & 48.6 & 75.4 & 45.5 & 56.2 & 73.5 & 65.0 & 35.1 & 60.1 & 55.7 \\
        \bf \our{} & 43.4 & 52.3 & 74.5 & 48.2 & 57.5 & 73.9 & 67.6 & 36.7 & 61.3 & 57.3 \\
        \bf \textcolor{black!60}{$\Delta$ \textit{compare to MoE-LLaVA}} & \textcolor{red}{+1.1$\!\!$} & \textcolor{red}{+3.7$\!\!$} & \textcolor{blue}{-0.9$\!\!$} & \textcolor{red}{+2.7$\!\!$} & \textcolor{red}{+1.3$\!\!$} & \textcolor{red}{+0.4$\!\!$} & \textcolor{red}{+2.6$\!\!$} & \textcolor{red}{+1.6$\!\!$} & \textcolor{red}{+1.2$\!\!$} & \textcolor{red}{+1.6$\!\!$} \\
        \bottomrule
        \end{tabular}
        }
    \end{center}
\end{table*}

\subsection{Main results} We compared the performance of ternary up-cycling \our{} to MoE-LLaVA across different model sizes on various multimodal tasks. As shown in Table~\ref{tab:main}, \our{} underperformed full-precision up-cycling MoE-LLaVA when converting a 0.5B dense model to an MoE model. However, the performance gap between \our{} and MoE-LLaVA narrows as the parameter counts of the dense model increases. Similar phenomenons are also reported by the low-bit pre-training of LLMs~\cite{bitnet, bitnetx, spectra}, which suggests promising trends of scaling model size for ternary MoEs. 

As the model size scales to 1.5B parameters, due to larger total parameter counts, \our{} surpasses MoE-LLaVA across various image understanding tasks, achieving an average accuracy improvement of 1.7\% with the same FLOPs. This demonstrates the effectiveness of our proposed method. Moreover, since the expert weights in \our{} are trained to adapt to ternary values, despite it has larger total parameter counts, the ternary MoE layer can be losslessly compressed to low-bit after training, significantly reducing the memory footprint caused by the ensemble of experts. As shown in Table~\ref{tab:compare}, at the 3B model size, \our{}'s expert memory is only 6.8GB — just 38\% of MoE-LLaVA’s 18.1GB.

\begin{table*}[t]
    \small
    \caption{The results of \our{} and MoE-LLaVA given the same amount of expert memory in 1.5B and 3B model size. Both of them are combined with post-training quantization (PTQ). The expert memory footprint includes contributions from both shared and routed experts.}
    \label{tab:ptq}
    \begin{center}
    \scalebox{0.92}{
        \begin{tabular}{l|c|cccccccccc}
        \toprule
        \bf Method & \bf Expert Memory$\downarrow$ & \makecell{\bf MMMU$\uparrow$ \\ (val)} & \makecell{\bf MMB$\uparrow$ \\ (en test)} & \makecell{\bf Seed$^{2+}$$\uparrow$ \\ (test)} & \makecell{\bf AI2D$\uparrow$ \\ (test)} & \makecell{\bf DocVQA$\uparrow$ \\ (val)} & \bf Avg.$\uparrow$ \\
        \midrule
        \multicolumn{6}{l}{\ \emph{1.5B Model Up-cycling}} \\
        \rowcolor{myblue} MoE-LLaVA + PTQ & 2.2GB & 41.1 & 68.0 & 53.1 & 67.3 & 55.0 & 56.9 \\
        \rowcolor{myblue} \our{} + PTQ & 2.2GB & 42.7 & 70.1 & 54.4 & 68.2 & 57.4 & 58.6 \\
        \rowcolor{mypink} MoE-LLaVA + PTQ & 1.6GB & 36.0 & 60.3 & 49.8 & 62.6 & 50.0 & 51.7\\
        \rowcolor{mypink} \our{} + PTQ & 1.6GB & 40.3 & 69.3 & 55.2 & 67.8 & 57.1 & 57.9 \\
        \midrule
        \multicolumn{6}{l}{\ \emph{3B Model Up-cycling}} \\
        \rowcolor{myblue} MoE-LLaVA + PTQ & 4.5GB & 42.2 & 75.3 & 55.4 & 72.3 & 59.4 & 60.9\\
        \rowcolor{myblue} \our{} + PTQ & 4.5GB & 43.2 & 74.8 & 57.0 & 73.3 & 60.9 & 61.8 \\
        \rowcolor{mypink} MoE-LLaVA + PTQ & 3.4GB & 37.7 & 69.7 & 52.2 & 67.5 & 56.8 & 56.8 \\
        \rowcolor{mypink} \our{} + PTQ & 3.4GB & 42.8 & 71.9 & 56.9 & 73.0 & 60.9 & 61.1 \\
        \bottomrule
        \end{tabular}
    }
    \end{center}
\end{table*}

\begin{table*}
\small
\caption{The results of \our{} and the other methods in similar model size on general VQA and multimodal reasoning tasks.}
\label{tab:sota}
\begin{center}
    \scalebox{0.85}{
    \begin{tabular}{l|c|ccccccc}
    \toprule
    \bf Model & \makecell{\bf Training \\ \bf Tokens} & \makecell{\bf MMMU \\ (val)} & \makecell{\bf MMB \\ (en test)} & \makecell{\bf Seed \\ (image)} & \makecell{\bf MMS \\ (test)} & \makecell{\bf MMV \\ (test)} & \makecell{\bf MathV \\ (testmini)} & \bf Avg.$\uparrow$ \\
    \midrule
    \multicolumn{6}{l}{\ \emph{Dense Model}} \\
    MM1.5-1B~\cite{mm1_5} & $>$200B & 35.8 & - & 70.2 & - & 37.4 & 37.2  & - \\
    MM1.5-3B~\cite{mm1_5} & $>$200B  & 37.1 & - & 72.4 & - & 41.0 & 44.4 & -\\
    MiniCPM-V2-3B~\cite{minicpmv2} & - & 38.2 & 69.1 & - & 41.7 & - & 38.7 & - \\
    TinyLLaVA-3B~\cite{tinyllava} & 4B & 39.9 & - & - & - & 34.8 & - & - \\
    Phi-3-Vision-4B~\cite{phi3} &  $>$0.8T & 40.4 & 73.9 & 71.8 & 47.9 & 45.4 & 44.5 & 54.0 \\
    Qwen2-VL-2B~\cite{qwen2vl} &  $>$1.4T & 41.1 & 74.9 & 72.1 & 48.0 & 49.5 & 43.0 & 54.8 \\
    \midrule
    \multicolumn{6}{l}{\ \emph{Sparse Model}} \\
    MoE-LLaVA~\cite{moe_llava} & 4B & 33.9 & 52.6 & 64.8 & 32.5 & 32.3 & 25.6 & 40.3 \\ 
    MolmoE-1B~\cite{molmo} & ~1.5B & 34.9 & 63.6 & 68.7 & 43.3 & 38.5 & 34.0 & 47.2 \\
    LLaVA-MoD-2B~\cite{llava_mod} & 10B & - & 68.9 & - & - & - & - & - \\
    MM1-3B-MoE~\cite{mm1} & $>$400B & 38.6 & 70.8 & 69.4 & - & 42.2 & 32.6 & - \\
    MM1-7B-MoE~\cite{mm1} & $>$400B & 40.9 & 72.7 & 70.9 & - & 45.2 & 40.9 & - \\
    MM1.5-1B-MoE~\cite{mm1_5} & $>$200B & 41.2 & - & 71.4 & - & 39.8 & 42.9 & - \\
    \midrule
    \rowcolor{myblue} \bf \our{}-1.5B (ours) & 21.6B & 40.4 & \bf 75.0 & \bf 72.5 & \bf 50.2 & \bf 52.6 & \bf 49.8 & \bf 56.8 \\ 
    \rowcolor{myblue} $\quad$ w/o initialize experts from FFN & 21.6B & \bf 41.8 & \bf 75.0 & 71.3 & 48.1 & 48.6 & 48.2 & 55.5 \\
    \bottomrule
    \end{tabular}
    }
\end{center}
\end{table*}

\subsection{Compatibility with post-training quantization} 
Despite the MoE layers of our model contain ternary experts, there still leaves a shared expert in full-precision in each layer. These shared experts can be quantized into low-bit using post-training quantization methods. 

We apply GPTQ~\cite{gptq} and AWQ~\cite{awq} at various bit-widths and report the best results given the same expert memory footprint. We use 512 samples with the length of 2048 tokens from Stage III's data as the calibration set. For MoE-LLaVA, all full-precision experts are quantized, resulting in expert memory footprints of 2.2GB and 4.5GB under INT4 quantization for the 1.5B and 3B models, respectively. To ensure a fair comparison, we quantize the shared expert of \our{} to INT8 using RTN~\cite{llmint8}. Additionally, we extend the comparison to scenarios with lower memory constraints. For expert memory footprints of 1.6GB and 3.4GB in the 1.5B and 3B models, MoE-LLaVA’s experts are quantized to 3-bit integers using GPTQ, while the shared experts of \our{} are quantized to INT4.

Table~\ref{tab:ptq} presents the results for \our{} and MoE-LLaVA, both combined with post-training quantization. Given the same expert memory footprint, \our{} achieves better performance than MoE-LLaVA. Under the same expert memory footprint, our method outperforms MoE-LLaVA across different model sizes. Notably, under stricter memory constraints, we observe a significant performance drop for MoE-LLaVA combined with GPTQ at 3-bit precision. However, since the parameters of our MoE layer are ternary, we can achieve the same memory footprint by applying INT4 quantization only to the shared expert. This further amplifies the advantages of our approach. Specifically, given the same expert memory of 3.4GB, \our{} achieves a gain of 4.3\% average accuracy compared with MoE-LLaVA on the end tasks. These results demonstrate that our method can achieve lower memory footprint combined with post-training quantization, while maintaining competitive performance.

\subsection{Scaling with more data}
\label{sec:sota}

To examine whether our method is friendly for scaling with data, we train a 1.5B \our{} model with more data during ternary up-cycling. We adopt the same data recipe for Stage I and Stage II as shown in Section~\ref{sec:setup}. Then we use a full set of MammoTH-VL~\cite{mammothvl} for Stage III, which contains 10 million samples, each associated with a single image. Every dense layer is replaced with an \our{} layer with one full-precision shared expert and four routed ternary experts. The training steps is set as 40k. The other hyper-parameters are consistent with the setup presented in Section~\ref{sec:setup}.

Table~\ref{tab:sota} summarizes the zero-shot accuracy of \our{} and the baselines across various multimodal reasoning and general VQA tasks. For the baselines, we use their reported scores when available; otherwise, we evaluate the open-sourced models using the same prompts as ours to ensure a fair comparison. As shown in Table~\ref{tab:sota}, although \our{}-1.5B is only trained with 21.6B tokens, our model achieves an improvement of 2.0\% average accuracy compared to Qwen2-VL-2B~\cite{qwen2vl}. Furthermore, \our{} outperforms the larger dense model with fewer FLOPs. Specifically, \our{} outperforms MiniCPM-V-2.0-3B and Phi-3-Vision-4B by a gain of 11.1\% and 5.3\% accuracy on the \textit{testmini} set of MathVista. 

For sparse model, due to stronger base LLM and vision encoder, our model significantly outperforms MoE-LLaVA of similar total and active model size by a gain of 16.5\% average accuracy. Notably, MM1.5-1B-MoE is a strong multimodal MoE baseline, which was trained from an 1B dense model with 64 experts replacing dense layers every two layers. \our{} outperforms it by a gain of 0.6\%, 1.1\%, 12.8\% and 6.9\% on MMMU, SeedBench (image), MMVet and MathVista, respectively. These results proves the effectiveness of the proposed \our{} on multimodal reasoning and general VQA.

\subsection{Ablation studies}
\label{sec:abl}

\begin{table*}[t]
    \centering
    \small
    \caption{Ablations on the precision of routed experts in \our{}.}
    \label{tab:bit_routed}
    \scalebox{1.0}{
    \begin{tabular}{c|ccccccc}
    \toprule
    \makecell{\bf Precision of \\ \bf Routed Expert} & \makecell{\bf MMMU \\ (val)} & \makecell{\bf MMB \\ (en test)} & \makecell{\bf AI2D \\ (test)} & \makecell{\bf ChartQA \\ (test)} & \makecell{\bf Seed$^{2+}$ \\ (test)} & \makecell{\bf MMS \\ (test)} &  \bf Avg.$\uparrow$ \\
    \midrule
    1-bit & 40.3 & 69.5 & 67.6 & 60.2 & 53.9 & 43.1 & 55.7 \\
    \rowcolor{myblue} \bf 1.58-bit & \bf 42.6 & \bf 70.0 & \bf 68.7 & \bf 61.3 & \bf 54.8 & \bf 46.4 & \bf 57.3 \\
    \bottomrule
    \end{tabular}
    }
\end{table*}

\begin{table*}[t]
    \centering
    \small
    \caption{Ablations on the precision of shared experts and the initialization methods of routed experts in \our{}.}
    \label{tab:bit_shard}
    \scalebox{1.0}{
    \begin{tabular}{c|c|ccccccc}
    \toprule
    \makecell{\bf Precision of \\ \bf Shared Expert} & \makecell{\bf Initialize \\
    \bf from FFN} & \makecell{\bf MMMU \\ (val)} & \makecell{\bf MMB \\ (en test)} & \makecell{\bf AI2D \\ (test)} & \makecell{\bf ChartQA \\ (test)} & \makecell{\bf Seed$^{2+}$ \\ (test)} & \makecell{\bf MMS \\ (test)} &  \bf Avg.$\uparrow$ \\
    \midrule
    Ternary & \xmark & 34.6 & 49.4 & 62.7 & 56.4 & 46.2 & 39.8 & 48.2 \\
     BF16 & \xmark & 40.1 & 69.9 & 67.1 & 59.9 & 53.2 & 44.5 & 55.8 \\
    \rowcolor{myblue} \bf BF16 & \cmark & \bf 42.6 & \bf 70.0 & \bf 68.7 & \bf 61.3 & \bf 54.8 & \bf 46.4 & \bf 57.3 \\
    \bottomrule
    \end{tabular}
    }
\end{table*}

\begin{table*}[t]
    \centering
    \small
    \caption{Ablations on the training recipe of \our{}. Given the same training FLOPs, we do not observe performance improvement from initially training with full-precision experts then fine-tuning them into ternary precision.}
    \label{tab:bit_train}
    \scalebox{1.0}{
    \begin{tabular}{c|c|ccccccc}
    \toprule
    \makecell{\bf Ternary \\ \bf Training} & \makecell{\bf Full-Precision \\ \bf Training} & \makecell{\bf MMMU \\ (val)} & \makecell{\bf MMB \\ (en test)} & \makecell{\bf AI2D \\ (test)} & \makecell{\bf ChartQA \\ (test)} & \makecell{\bf Seed$^{2+}$ \\ (test)} & \makecell{\bf MMS \\ (test)} &  \bf Avg.$\uparrow$ \\
    \midrule
    20\%  & 80\% & 39.3 & 60.5 & 62.6 & 56.8 & 53.2 & 42.0 & 52.4 \\
    60\%  & 40\% & 41.3 & 64.0 & 65.3 & 57.0 & 54.0 & 45.1 & 54.4 \\
    \rowcolor{myblue} \bf 100\% & \bf 0\%  & \bf 42.6 & \bf 70.0 & \bf 68.7 & \bf 61.3 & \bf 54.8 & \bf 46.4 & \bf 57.3 \\
    \bottomrule
    \end{tabular}
    }
\end{table*}

\paragraph{Precision of routed experts.} We investigate the impact of expert precision on the performance of \our{}. 
Specifically, we compare ternary (i.e., 1.58-bit) up-cycling to 1-bit up-cycling with BWN~\cite{xnor} as the weight quantizers.
Both models are up-cycled from Qwen2.5-1.5B with SigLIP-L as the vision encoder to ensure a fair comparison. 
As shown in Table~\ref{tab:bit_routed}, using binary experts results in performance degradation across most tasks.
Similar findings have been reported in the quantization-aware training of BERT models~\cite{binarybert}, where transitioning from ternary to binary weights leads to a substantially more complex and irregular loss landscape, making optimization notably more difficult. 
Above all, ternary up-cycling is a memory-effective and high-performance solution for MoE models.

\paragraph{Precision of shared experts.} We ablate the effect of the precision of the shared expert reused from the FFN of pre-trained dense checkpoint. \our{} retains the precision of shared expert as BF16 and freezes the modules during up-cycling. We compare it to a model with the ternary shared expert. All ternary experts are trainable. Table~\ref{tab:bit_shard} presents the zero-shot performance of these models on MMMU, MMBench, AI2D, ChartQA, SeedBench-2-Plus and MMStar tasks. Weight ternarization of the shared experts has significant effect on overall performance. Specifically, the model with full-precison shared experts outperforms it with ternary shared experts by an improvement of 7.6\% average accuracy on the end tasks. This demonstrates the importance of keeping the pre-trained FFN as a high-precision shared expert during ternary up-cycling.

\paragraph{Initialization of routed experts.} We compare \our{} to randomly initialized routed experts in Stage III. Table~\ref{tab:bit_shard} presents the results for a 1.5B model, where initializing from the FFN yields a 1.5\% improvement in average accuracy on end tasks compared to random initialization. Moreover, we analyze the impact of data scaling using the data recipe described in Section~\ref{sec:sota}. As demonstrated in Table~\ref{tab:sota}, FFN-based initialization maintains its advantage with additional training data, achieving a 1.3\% higher average accuracy than random initialization. These findings suggest that leveraging a pre-trained full-precision FFN for \our{}'s initialization not only enhances performance but also accelerates the convergence of ternary experts. Additional results for the 0.5B and 3B models are provided in the Appendix~\ref{ap:abl}.

\paragraph{Training recipe.} We conduct ablation studies on the training strategy of ternary up-cycling in \our{} to assess the effectiveness of first training with full-precision experts before fine-tuning the model to ternary precision. All models are trained on 6.25B tokens and up-cycled from Qwen2.5-1.5B. We vary the proportion of training conducted in full-precision versus ternary precision. As shown in Table~\ref{tab:bit_train}, we do not observe performance gain from initially training with full-precision experts. In fact, accuracy improves as the proportion of ternary training increases. Therefore, for both simplicity and improved performance, \our{} is trained directly in ternary precision without a full-precision training phase during up-cycling.

\begin{figure*}[t]
    \centering
    \begin{subfigure}{0.325\textwidth}
        \centering
        \includegraphics[width=\textwidth]{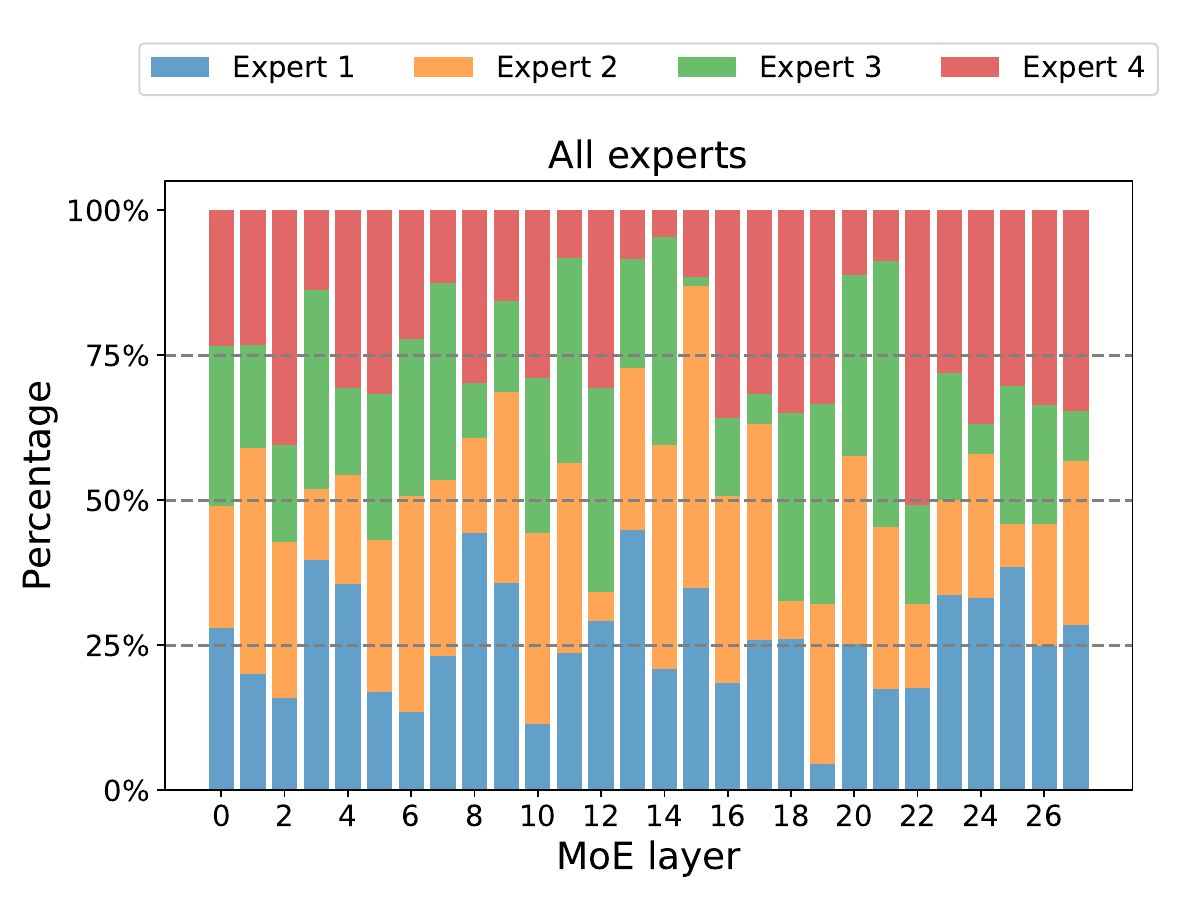}
        \caption{All tokens.}
        \label{fig:mmbench_en_test:all}
    \end{subfigure}
    \begin{subfigure}{0.325\textwidth}
        \centering
        \includegraphics[width=\textwidth]{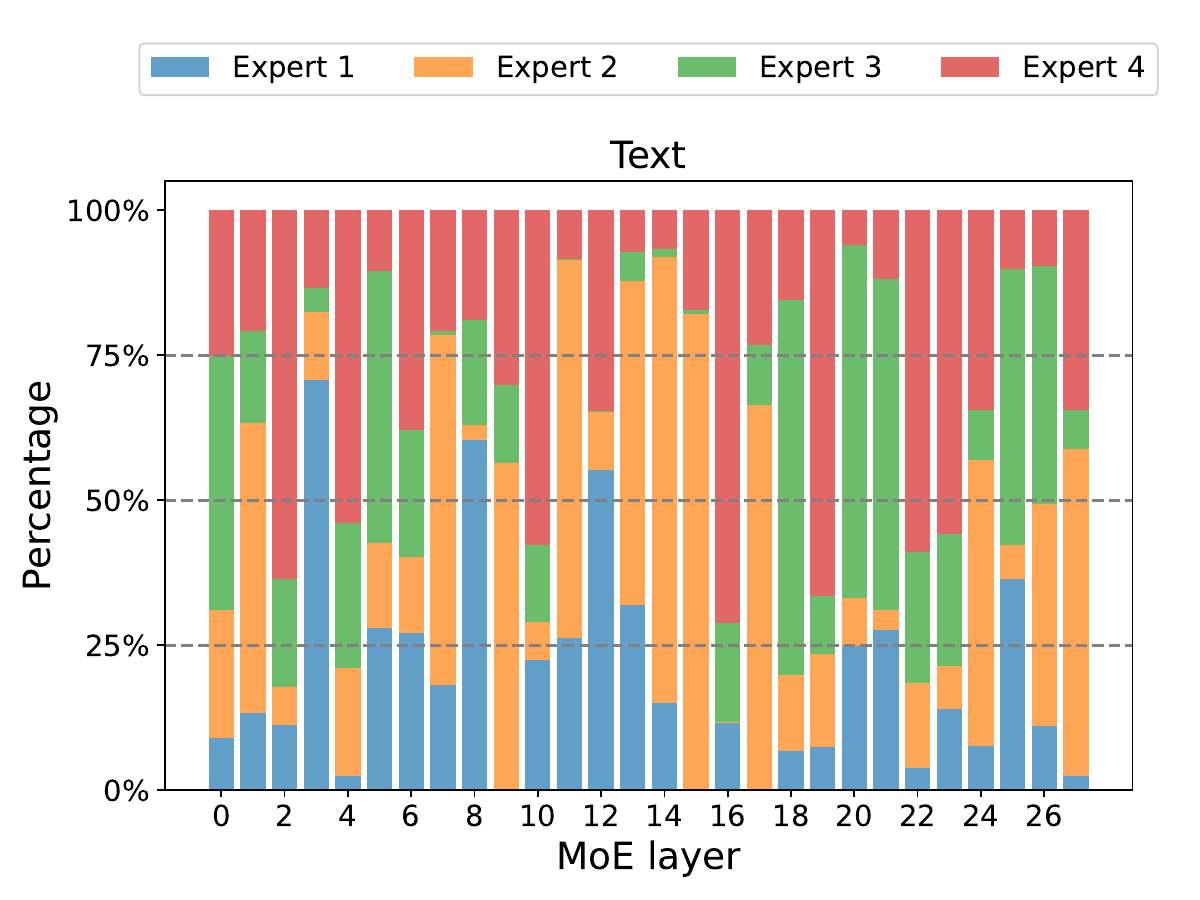}
        \caption{Text tokens.}
        \label{fig:mmbench_en_test:text}
    \end{subfigure}
    \begin{subfigure}{0.325\textwidth}
        \centering
        \includegraphics[width=\textwidth]{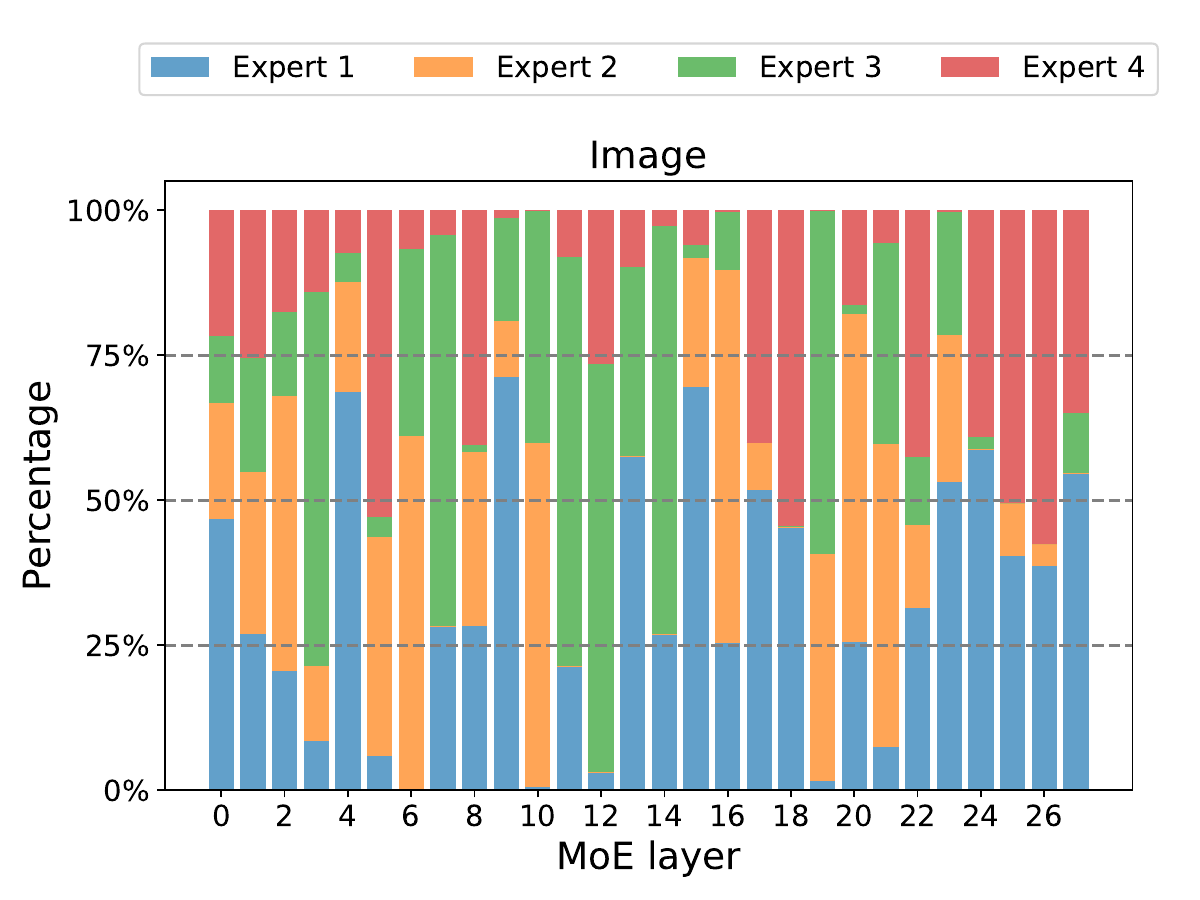}
        \caption{Image tokens.}
        \label{fig:mmbench_en_test:img}
    \end{subfigure}
    \caption{Visualization of the routing distributions of all tokens, text tokens, image tokens across all experts on the \textit{en-test} set of MMBench.}
    \label{fig:mmbench_en_test:router_dist1}
\end{figure*}

\section{Analysis}

We visualize the routing distribution of all tokens in \our{}-1.5B on the \textit{en-test} split of the MMBench dataset. As shown in Figure~\ref{fig:mmbench_en_test:all}, expert utilization across all tokens is well-balanced. To further investigate modality-specific behavior, we present the routing distributions for text and image tokens separately in Figures~\ref{fig:mmbench_en_test:text} and \ref{fig:mmbench_en_test:img}, respectively. Notably, text and image tokens exhibit distinct routing patterns. For example, expert \#1 is frequently activated for image tokens in the first layer and the final five layers. 
Additional visualizations across various tasks are provided in Appendix~\ref{ap:vis:rt}. We observe that routing distributions remain largely consistent across different tasks, suggesting that the experts in \our{} specialize based on modality rather than task-specific features. Moreover, we include per-expert routing distributions by modality in Appendix~\ref{ap:vis:expert}. Interestingly, some experts exhibit clear modality preferences despite the absence of explicit modality conditioning during training. 
To better understand expert specialization, we further apply PCA~\cite{pca} to extract the top-10 routing pathways for text and image tokens. More visualizations are included in Appendix\ref{ap:vis:path}. These findings enhance our understanding of \our{}'s behavior and workflow from a token-level perspective.

\section{Conclusion}
In this work, we introduce \our{}, a scalable and memory-efficient approach to train multimodal Mixture-of-Ternary-Experts models from full-precision dense checkpoints. Extensive experiments show that our model matches the full-precision up-cycling MoE-LLaVA in zero-shot performance on end tasks, starting from model sizes exceeding 1.5B parameters. Furthermore, \our{} is compatible with post-training quantization methods, enabling further reductions in the memory footprint of MoE models. Given the same expert memory footprint of 3.4GB, \our{} surpasses MoE-LLaVA with an average accuracy gain of 4.3\% on image understanding tasks, highlighting the effectiveness of our approach, particularly for memory-constrained edge devices.

\bibliography{mote}

@article{moe_llava,
  author       = {Bin Lin and
                  Zhenyu Tang and
                  Yang Ye and
                  Jiaxi Cui and
                  Bin Zhu and
                  Peng Jin and
                  Junwu Zhang and
                  Munan Ning and
                  Li Yuan},
  title        = {MoE-LLaVA: Mixture of Experts for Large Vision-Language Models},
  journal      = {CoRR},
  volume       = {abs/2401.15947},
  year         = {2024},
}

@article{minicpmv2,
  author       = {Yuan Yao and
                  Tianyu Yu and
                  Ao Zhang and
                  Chongyi Wang and
                  Junbo Cui and
                  Hongji Zhu and
                  Tianchi Cai and
                  Haoyu Li and
                  Weilin Zhao and
                  Zhihui He and
                  Qianyu Chen and
                  Huarong Zhou and
                  Zhensheng Zou and
                  Haoye Zhang and
                  Shengding Hu and
                  Zhi Zheng and
                  Jie Zhou and
                  Jie Cai and
                  Xu Han and
                  Guoyang Zeng and
                  Dahai Li and
                  Zhiyuan Liu and
                  Maosong Sun},
  title        = {MiniCPM-V: {A} {GPT-4V} Level {MLLM} on Your Phone},
  journal      = {CoRR},
  volume       = {abs/2408.01800},
  year         = {2024},
}

@article{qwen2vl,
  title={Qwen2-VL: Enhancing Vision-Language Model's Perception of the World at Any Resolution},
  author={Wang, Peng and Bai, Shuai and Tan, Sinan and Wang, Shijie and Fan, Zhihao and Bai, Jinze and Chen, Keqin and Liu, Xuejing and Wang, Jialin and Ge, Wenbin and Fan, Yang and Dang, Kai and Du, Mengfei and Ren, Xuancheng and Men, Rui and Liu, Dayiheng and Zhou, Chang and Zhou, Jingren and Lin, Junyang},
  journal={arXiv preprint arXiv:2409.12191},
  year={2024}
}

@inproceedings{mm1,
  author       = {Brandon McKinzie and
                  Zhe Gan and
                  Jean{-}Philippe Fauconnier and
                  Sam Dodge and
                  Bowen Zhang and
                  Philipp Dufter and
                  Dhruti Shah and
                  Xianzhi Du and
                  Futang Peng and
                  Anton Belyi and
                  Haotian Zhang and
                  Karanjeet Singh and
                  Doug Kang and
                  Hongyu H{\`{e}} and
                  Max Schwarzer and
                  Tom Gunter and
                  Xiang Kong and
                  Aonan Zhang and
                  Jianyu Wang and
                  Chong Wang and
                  Nan Du and
                  Tao Lei and
                  Sam Wiseman and
                  Mark Lee and
                  Zirui Wang and
                  Ruoming Pang and
                  Peter Grasch and
                  Alexander Toshev and
                  Yinfei Yang},
  title        = {{MM1:} Methods, Analysis and Insights from Multimodal {LLM} Pre-training},
  booktitle    = {{ECCV} 2024},
  volume       = {15087},
  pages        = {304--323},
  year         = {2024},
}

@article{mm1_5,
  author       = {Haotian Zhang and
                  Mingfei Gao and
                  Zhe Gan and
                  Philipp Dufter and
                  Nina Wenzel and
                  Forrest Huang and
                  Dhruti Shah and
                  Xianzhi Du and
                  Bowen Zhang and
                  Yanghao Li and
                  Sam Dodge and
                  Keen You and
                  Zhen Yang and
                  Aleksei Timofeev and
                  Mingze Xu and
                  Hong{-}You Chen and
                  Jean{-}Philippe Fauconnier and
                  Zhengfeng Lai and
                  Haoxuan You and
                  Zirui Wang and
                  Afshin Dehghan and
                  Peter Grasch and
                  Yinfei Yang},
  title        = {{MM1.5:} Methods, Analysis {\&} Insights from Multimodal {LLM}
                  Fine-tuning},
  journal      = {CoRR},
  volume       = {abs/2409.20566},
  year         = {2024},
}

@article{internvl2,
  author       = {Zhe Chen and
                  Weiyun Wang and
                  Hao Tian and
                  Shenglong Ye and
                  Zhangwei Gao and
                  Erfei Cui and
                  Wenwen Tong and
                  Kongzhi Hu and
                  Jiapeng Luo and
                  Zheng Ma and
                  Ji Ma and
                  Jiaqi Wang and
                  Xiaoyi Dong and
                  Hang Yan and
                  Hewei Guo and
                  Conghui He and
                  Botian Shi and
                  Zhenjiang Jin and
                  Chao Xu and
                  Bin Wang and
                  Xingjian Wei and
                  Wei Li and
                  Wenjian Zhang and
                  Bo Zhang and
                  Pinlong Cai and
                  Licheng Wen and
                  Xiangchao Yan and
                  Min Dou and
                  Lewei Lu and
                  Xizhou Zhu and
                  Tong Lu and
                  Dahua Lin and
                  Yu Qiao and
                  Jifeng Dai and
                  Wenhai Wang},
  title        = {How Far Are We to GPT-4V? Closing the Gap to Commercial Multimodal
                  Models with Open-Source Suites},
  journal      = {CoRR},
  volume       = {abs/2404.16821},
  year         = {2024},
}

@article{lmm-eval,
  title={Lmms-eval: Reality check on the evaluation of large multimodal models},
  author={Zhang, Kaichen and Li, Bo and Zhang, Peiyuan and Pu, Fanyi and Cahyono, Joshua Adrian and Hu, Kairui and Liu, Shuai and Zhang, Yuanhan and Yang, Jingkang and Li, Chunyuan and others},
  journal={arXiv preprint arXiv:2407.12772},
  year={2024}
}

@article{molmo,
  title={Molmo and pixmo: Open weights and open data for state-of-the-art multimodal models},
  author={Deitke, Matt and Clark, Christopher and Lee, Sangho and Tripathi, Rohun and Yang, Yue and Park, Jae Sung and Salehi, Mohammadreza and Muennighoff, Niklas and Lo, Kyle and Soldaini, Luca and others},
  journal={arXiv preprint arXiv:2409.17146},
  year={2024}
}

@inproceedings{upcycling,
  author       = {Aran Komatsuzaki and
                  Joan Puigcerver and
                  James Lee{-}Thorp and
                  Carlos Riquelme Ruiz and
                  Basil Mustafa and
                  Joshua Ainslie and
                  Yi Tay and
                  Mostafa Dehghani and
                  Neil Houlsby},
  title        = {Sparse Upcycling: Training Mixture-of-Experts from Dense Checkpoints},
  booktitle    = {{ICLR} 2023},
  publisher    = {OpenReview.net},
  year         = {2023},
}

@inproceedings{knowledge_neuron,
  author       = {Damai Dai and
                  Li Dong and
                  Yaru Hao and
                  Zhifang Sui and
                  Baobao Chang and
                  Furu Wei},
  title        = {Knowledge Neurons in Pretrained Transformers},
  booktitle    = {{ACL} 2022},
  pages        = {8493--8502},
  year         = {2022},
}

@inproceedings{tran_are_kv,
  author       = {Mor Geva and
                  Roei Schuster and
                  Jonathan Berant and
                  Omer Levy},
  title        = {Transformer Feed-Forward Layers Are Key-Value Memories},
  booktitle    = {{EMNLP} 2021},
  pages        = {5484--5495},
  year         = {2021},
}

@inproceedings{gshard,
  author       = {Dmitry Lepikhin and
                  HyoukJoong Lee and
                  Yuanzhong Xu and
                  Dehao Chen and
                  Orhan Firat and
                  Yanping Huang and
                  Maxim Krikun and
                  Noam Shazeer and
                  Zhifeng Chen},
  title        = {GShard: Scaling Giant Models with Conditional Computation and Automatic
                  Sharding},
  booktitle    = {{ICLR} 2021,},
  year         = {2021},
}

@article{bitnetx,
  author       = {Shuming Ma and
                  Hongyu Wang and
                  Lingxiao Ma and
                  Lei Wang and
                  Wenhui Wang and
                  Shaohan Huang and
                  Li Dong and
                  Ruiping Wang and
                  Jilong Xue and
                  Furu Wei},
  title        = {The Era of 1-bit LLMs: All Large Language Models are in 1.58 Bits},
  journal      = {CoRR},
  volume       = {abs/2402.17764},
  year         = {2024},
}

@article{switch,
  author       = {William Fedus and
                  Barret Zoph and
                  Noam Shazeer},
  title        = {Switch Transformers: Scaling to Trillion Parameter Models with Simple
                  and Efficient Sparsity},
  journal      = {J. Mach. Learn. Res.},
  volume       = {23},
  pages        = {120:1--120:39},
  year         = {2022},
}

@article{ste,
  author       = {Yoshua Bengio and
                  Nicholas L{\'{e}}onard and
                  Aaron C. Courville},
  title        = {Estimating or Propagating Gradients Through Stochastic Neurons for
                  Conditional Computation},
  journal      = {CoRR},
  volume       = {abs/1308.3432},
  year         = {2013},
}

@article{mimic,
  author       = {Bo Li and
                  Yuanhan Zhang and
                  Liangyu Chen and
                  Jinghao Wang and
                  Fanyi Pu and
                  Jingkang Yang and
                  Chunyuan Li and
                  Ziwei Liu},
  title        = {{MIMIC-IT:} Multi-Modal In-Context Instruction Tuning},
  journal      = {CoRR},
  volume       = {abs/2306.05425},
  year         = {2023},
}

@inproceedings{lrv,
  title={Mitigating hallucination in large multi-modal models via robust instruction tuning},
  author={Liu, Fuxiao and Lin, Kevin and Li, Linjie and Wang, Jianfeng and Yacoob, Yaser and Wang, Lijuan},
  booktitle={The Twelfth International Conference on Learning Representations},
  year={2023}
}

@article{svit,
  author       = {Bo Zhao and
                  Boya Wu and
                  Tiejun Huang},
  title        = {{SVIT:} Scaling up Visual Instruction Tuning},
  journal      = {CoRR},
  volume       = {abs/2307.04087},
  year         = {2023},
}

@article{lvis,
  title={To see is to believe: Prompting gpt-4v for better visual instruction tuning},
  author={Wang, Junke and Meng, Lingchen and Weng, Zejia and He, Bo and Wu, Zuxuan and Jiang, Yu-Gang},
  journal={arXiv preprint arXiv:2311.07574},
  year={2023}
}

@inproceedings{llava_1_5,
  title={Improved baselines with visual instruction tuning},
  author={Liu, Haotian and Li, Chunyuan and Li, Yuheng and Lee, Yong Jae},
  booktitle={Proceedings of the IEEE/CVF Conference on Computer Vision and Pattern Recognition},
  pages={26296--26306},
  year={2024}
}

@article{mammothvl,
      title={MAmmoTH-VL: Eliciting Multimodal Reasoning with Instruction Tuning at Scale}, 
      author={Jarvis Guo and Tuney Zheng and Yuelin Bai and Bo Li and Yubo Wang and King Zhu and Yizhi Li and Graham Neubig and Wenhu Chen and Xiang Yue},
      year={2024},
      eprint={2412.05237},
      archivePrefix={arXiv},
      primaryClass={cs.CL},
      url={https://arxiv.org/abs/2412.05237}, 
}

@inproceedings {ladder,
    author = {Lei Wang and Lingxiao Ma and Shijie Cao and Quanlu Zhang and Jilong Xue and Yining Shi and Ningxin Zheng and Ziming Miao and Fan Yang and Ting Cao and Yuqing Yang and Mao Yang},
    title = {Ladder: Enabling Efficient Low-Precision Deep Learning Computing through Hardware-aware Tensor Transformation},
    booktitle = {18th USENIX Symposium on Operating Systems Design and Implementation (OSDI 24)},
    year = {2024},
}

@inproceedings{siglip,
  title={Sigmoid loss for language image pre-training},
  author={Zhai, Xiaohua and Mustafa, Basil and Kolesnikov, Alexander and Beyer, Lucas},
  booktitle={Proceedings of the IEEE/CVF International Conference on Computer Vision},
  pages={11975--11986},
  year={2023}
}

@article{qwen2_5,
  title={Qwen2. 5 technical report},
  author={Yang, An and Yang, Baosong and Zhang, Beichen and Hui, Binyuan and Zheng, Bo and Yu, Bowen and Li, Chengyuan and Liu, Dayiheng and Huang, Fei and Wei, Haoran and others},
  journal={arXiv preprint arXiv:2412.15115},
  year={2024}
}

@article{bitnet,
  title={Bitnet: Scaling 1-bit transformers for large language models},
  author={Wang, Hongyu and Ma, Shuming and Dong, Li and Huang, Shaohan and Wang, Huaijie and Ma, Lingxiao and Yang, Fan and Wang, Ruiping and Wu, Yi and Wei, Furu},
  journal={arXiv preprint arXiv:2310.11453},
  year={2023}
}

@article{spectra,
  title={Spectra: Surprising effectiveness of pretraining ternary language models at scale},
  author={Kaushal, Ayush and Vaidhya, Tejas and Mondal, Arnab Kumar and Pandey, Tejas and Bhagat, Aaryan and Rish, Irina},
  journal={arXiv preprint arXiv:2407.12327},
  year={2024}
}

@article{gptq,
  title={Gptq: Accurate post-training quantization for generative pre-trained transformers},
  author={Frantar, Elias and Ashkboos, Saleh and Hoefler, Torsten and Alistarh, Dan},
  journal={arXiv preprint arXiv:2210.17323},
  year={2022}
}

@article{awq,
  title={AWQ: Activation-aware Weight Quantization for On-Device LLM Compression and Acceleration},
  author={Lin, Ji and Tang, Jiaming and Tang, Haotian and Yang, Shang and Chen, Wei-Ming and Wang, Wei-Chen and Xiao, Guangxuan and Dang, Xingyu and Gan, Chuang and Han, Song},
  journal={Proceedings of Machine Learning and Systems},
  volume={6},
  pages={87--100},
  year={2024}
}

@article{llmint8,
  author       = {Tim Dettmers and
                  Mike Lewis and
                  Younes Belkada and
                  Luke Zettlemoyer},
  title        = {LLM.int8(): 8-bit Matrix Multiplication for Transformers at Scale},
  journal      = {CoRR},
  volume       = {abs/2208.07339},
  year         = {2022},
}

@inproceedings{mmmu,
  title={Mmmu: A massive multi-discipline multimodal understanding and reasoning benchmark for expert agi},
  author={Yue, Xiang and Ni, Yuansheng and Zhang, Kai and Zheng, Tianyu and Liu, Ruoqi and Zhang, Ge and Stevens, Samuel and Jiang, Dongfu and Ren, Weiming and Sun, Yuxuan and others},
  booktitle={Proceedings of the IEEE/CVF Conference on Computer Vision and Pattern Recognition},
  pages={9556--9567},
  year={2024}
}

@inproceedings{mathvista,
  author    = {Lu, Pan and Bansal, Hritik and Xia, Tony and Liu, Jiacheng and Li, Chunyuan and Hajishirzi, Hannaneh and Cheng, Hao and Chang, Kai-Wei and Galley, Michel and Gao, Jianfeng},
  title     = {MathVista: Evaluating Mathematical Reasoning of Foundation Models in Visual Contexts},
  booktitle={International Conference on Learning Representations (ICLR)},
  year      = {2024}
}

@article{mmstar,
  title={Are We on the Right Way for Evaluating Large Vision-Language Models?},
  author={Chen, Lin and Li, Jinsong and Dong, Xiaoyi and Zhang, Pan and Zang, Yuhang and Chen, Zehui and Duan, Haodong and Wang, Jiaqi and Qiao, Yu and Lin, Dahua and others},
  journal={arXiv preprint arXiv:2403.20330},
  year={2024}
}

@inproceedings{mmbench,
  title={Mmbench: Is your multi-modal model an all-around player?},
  author={Liu, Yuan and Duan, Haodong and Zhang, Yuanhan and Li, Bo and Zhang, Songyang and Zhao, Wangbo and Yuan, Yike and Wang, Jiaqi and He, Conghui and Liu, Ziwei and others},
  booktitle={European conference on computer vision},
  pages={216--233},
  year={2024},
  organization={Springer}
}

@article{seedplus,
  title={Seed-bench-2-plus: Benchmarking multimodal large language models with text-rich visual comprehension},
  author={Li, Bohao and Ge, Yuying and Chen, Yi and Ge, Yixiao and Zhang, Ruimao and Shan, Ying},
  journal={arXiv preprint arXiv:2404.16790},
  year={2024}
}

@inproceedings{ai2d,
  title={A diagram is worth a dozen images},
  author={Kembhavi, Aniruddha and Salvato, Mike and Kolve, Eric and Seo, Minjoon and Hajishirzi, Hannaneh and Farhadi, Ali},
  booktitle={Computer Vision--ECCV 2016: 14th European Conference, Amsterdam, The Netherlands, October 11--14, 2016, Proceedings, Part IV 14},
  pages={235--251},
  year={2016},
  organization={Springer}
}

@article{chartqa,
  title={Chartqa: A benchmark for question answering about charts with visual and logical reasoning},
  author={Masry, Ahmed and Long, Do Xuan and Tan, Jia Qing and Joty, Shafiq and Hoque, Enamul},
  journal={arXiv preprint arXiv:2203.10244},
  year={2022}
}

@inproceedings{infovqa,
  title={Infographicvqa},
  author={Mathew, Minesh and Bagal, Viraj and Tito, Rub{\`e}n and Karatzas, Dimosthenis and Valveny, Ernest and Jawahar, CV},
  booktitle={Proceedings of the IEEE/CVF Winter Conference on Applications of Computer Vision},
  pages={1697--1706},
  year={2022}
}

@inproceedings{docvqa,
  title={Docvqa: A dataset for vqa on document images},
  author={Mathew, Minesh and Karatzas, Dimosthenis and Jawahar, CV},
  booktitle={Proceedings of the IEEE/CVF winter conference on applications of computer vision},
  pages={2200--2209},
  year={2021}
}

@article{mmvet,
  title={Mm-vet: Evaluating large multimodal models for integrated capabilities},
  author={Yu, Weihao and Yang, Zhengyuan and Li, Linjie and Wang, Jianfeng and Lin, Kevin and Liu, Zicheng and Wang, Xinchao and Wang, Lijuan},
  journal={arXiv preprint arXiv:2308.02490},
  year={2023}
}

@article{fp4,
  title={Optimizing Large Language Model Training Using FP4 Quantization},
  author={Wang, Ruizhe and Gong, Yeyun and Liu, Xiao and Zhao, Guoshuai and Yang, Ziyue and Guo, Baining and Zha, Zhengjun and Cheng, Peng},
  journal={arXiv preprint arXiv:2501.17116},
  year={2025}
}

@article{fp8,
  title={Fp8-lm: Training fp8 large language models},
  author={Peng, Houwen and Wu, Kan and Wei, Yixuan and Zhao, Guoshuai and Yang, Yuxiang and Liu, Ze and Xiong, Yifan and Yang, Ziyue and Ni, Bolin and Hu, Jingcheng and others},
  journal={arXiv preprint arXiv:2310.18313},
  year={2023}
}

@article{quip,
  title={Quip: 2-bit quantization of large language models with guarantees},
  author={Chee, Jerry and Cai, Yaohui and Kuleshov, Volodymyr and De Sa, Christopher M},
  journal={Advances in Neural Information Processing Systems},
  volume={36},
  year={2024}
}

@inproceedings{quip_sharp,
  author       = {Albert Tseng and
                  Jerry Chee and
                  Qingyao Sun and
                  Volodymyr Kuleshov and
                  Christopher De Sa},
  title        = {QuIP{\#}: Even Better {LLM} Quantization with Hadamard Incoherence
                  and Lattice Codebooks},
  booktitle    = {{ICML}},
  year         = {2024},
}

@article{qtip,
  title={QTIP: Quantization with Trellises and Incoherence Processing},
  author={Tseng, Albert and Sun, Qingyao and Hou, David and De Sa, Christopher},
  journal={arXiv preprint arXiv:2406.11235},
  year={2024}
}

@article{matmul_free,
  title={Scalable MatMul-free Language Modeling},
  author={Zhu, Rui-Jie and Zhang, Yu and Sifferman, Ethan and Sheaves, Tyler and Wang, Yiqiao and Richmond, Dustin and Zhou, Peng and Eshraghian, Jason K},
  journal={arXiv preprint arXiv:2406.02528},
  year={2024}
}

@article{deepseekv3,
  author       = {DeepSeek{-}AI and
                  Aixin Liu and
                  Bei Feng and
                  Bing Xue and
                  Bingxuan Wang and
                  Bochao Wu and
                  Chengda Lu and
                  Chenggang Zhao and
                  Chengqi Deng and
                  Chenyu Zhang and
                  Chong Ruan and
                  Damai Dai and
                  Daya Guo and
                  Dejian Yang and
                  Deli Chen and
                  Dongjie Ji and
                  Erhang Li and
                  Fangyun Lin and
                  Fucong Dai and
                  Fuli Luo and
                  Guangbo Hao and
                  Guanting Chen and
                  Guowei Li and
                  H. Zhang and
                  Han Bao and
                  Hanwei Xu and
                  Haocheng Wang and
                  Haowei Zhang and
                  Honghui Ding and
                  Huajian Xin and
                  Huazuo Gao and
                  Hui Li and
                  Hui Qu and
                  J. L. Cai and
                  Jian Liang and
                  Jianzhong Guo and
                  Jiaqi Ni and
                  Jiashi Li and
                  Jiawei Wang and
                  Jin Chen and
                  Jingchang Chen and
                  Jingyang Yuan and
                  Junjie Qiu and
                  Junlong Li and
                  Junxiao Song and
                  Kai Dong and
                  Kai Hu and
                  Kaige Gao and
                  Kang Guan and
                  Kexin Huang and
                  Kuai Yu and
                  Lean Wang and
                  Lecong Zhang and
                  Lei Xu and
                  Leyi Xia and
                  Liang Zhao and
                  Litong Wang and
                  Liyue Zhang and
                  Meng Li and
                  Miaojun Wang and
                  Mingchuan Zhang and
                  Minghua Zhang and
                  Minghui Tang and
                  Mingming Li and
                  Ning Tian and
                  Panpan Huang and
                  Peiyi Wang and
                  Peng Zhang and
                  Qiancheng Wang and
                  Qihao Zhu and
                  Qinyu Chen and
                  Qiushi Du and
                  R. J. Chen and
                  R. L. Jin and
                  Ruiqi Ge and
                  Ruisong Zhang and
                  Ruizhe Pan and
                  Runji Wang and
                  Runxin Xu and
                  Ruoyu Zhang and
                  Ruyi Chen and
                  S. S. Li and
                  Shanghao Lu and
                  Shangyan Zhou and
                  Shanhuang Chen and
                  Shaoqing Wu and
                  Shengfeng Ye and
                  Shengfeng Ye and
                  Shirong Ma and
                  Shiyu Wang and
                  Shuang Zhou and
                  Shuiping Yu and
                  Shunfeng Zhou and
                  Shuting Pan and
                  T. Wang and
                  Tao Yun and
                  Tian Pei and
                  Tianyu Sun and
                  W. L. Xiao and
                  Wangding Zeng},
  title        = {DeepSeek-V3 Technical Report},
  journal      = {CoRR},
  volume       = {abs/2412.19437},
  year         = {2024},
}

@article{unimoe,
  title={Uni-moe: Scaling unified multimodal llms with mixture of experts},
  author={Li, Yunxin and Jiang, Shenyuan and Hu, Baotian and Wang, Longyue and Zhong, Wanqi and Luo, Wenhan and Ma, Lin and Zhang, Min},
  journal={IEEE Transactions on Pattern Analysis and Machine Intelligence},
  year={2025},
  publisher={IEEE}
}

@article{olmoe,
  title={Olmoe: Open mixture-of-experts language models},
  author={Muennighoff, Niklas and Soldaini, Luca and Groeneveld, Dirk and Lo, Kyle and Morrison, Jacob and Min, Sewon and Shi, Weijia and Walsh, Pete and Tafjord, Oyvind and Lambert, Nathan and others},
  journal={arXiv preprint arXiv:2409.02060},
  year={2024}
}

@article{llava_mod,
  title={Llava-mod: Making llava tiny via moe knowledge distillation},
  author={Shu, Fangxun and Liao, Yue and Zhuo, Le and Xu, Chenning and Zhang, Lei and Zhang, Guanghao and Shi, Haonan and Chen, Long and Zhong, Tao and He, Wanggui and others},
  journal={arXiv preprint arXiv:2408.15881},
  year={2024}
}

@article{deepseekvl2,
  title={Deepseek-vl2: Mixture-of-experts vision-language models for advanced multimodal understanding},
  author={Wu, Zhiyu and Chen, Xiaokang and Pan, Zizheng and Liu, Xingchao and Liu, Wen and Dai, Damai and Gao, Huazuo and Ma, Yiyang and Wu, Chengyue and Wang, Bingxuan and others},
  journal={arXiv preprint arXiv:2412.10302},
  year={2024}
}

@article{phi3,
  title={Phi-3 technical report: A highly capable language model locally on your phone},
  author={Abdin, Marah and Aneja, Jyoti and Awadalla, Hany and Awadallah, Ahmed and Awan, Ammar Ahmad and Bach, Nguyen and Bahree, Amit and Bakhtiari, Arash and Bao, Jianmin and Behl, Harkirat and others},
  journal={arXiv preprint arXiv:2404.14219},
  year={2024}
}

@misc{qwen2_5_vl,
      title={Qwen2.5-VL Technical Report}, 
      author={Shuai Bai and Keqin Chen and Xuejing Liu and Jialin Wang and Wenbin Ge and Sibo Song and Kai Dang and Peng Wang and Shijie Wang and Jun Tang and Humen Zhong and Yuanzhi Zhu and Mingkun Yang and Zhaohai Li and Jianqiang Wan and Pengfei Wang and Wei Ding and Zheren Fu and Yiheng Xu and Jiabo Ye and Xi Zhang and Tianbao Xie and Zesen Cheng and Hang Zhang and Zhibo Yang and Haiyang Xu and Junyang Lin},
      year={2025},
      eprint={2502.13923},
      archivePrefix={arXiv},
      primaryClass={cs.CV},
}

@article{glu,
  title={Glu variants improve transformer},
  author={Shazeer, Noam},
  journal={arXiv preprint arXiv:2002.05202},
  year={2020}
}

@article{bitnetcpp,
  title={1-bit AI Infra: Part 1.1, Fast and Lossless BitNet b1. 58 Inference on CPUs},
  author={Wang, Jinheng and Zhou, Hansong and Song, Ting and Mao, Shaoguang and Ma, Shuming and Wang, Hongyu and Xia, Yan and Wei, Furu},
  journal={arXiv preprint arXiv:2410.16144},
  year={2024}
}

@article{pca,
    author = {Karl Pearson},
    title = {LIII. On lines and planes of closest fit to systems of points in space },
    journal = {The London, Edinburgh, and Dublin Philosophical Magazine and Journal of Science},
    volume = {2},
    number = {11},
    pages = {559--572},
    year = {1901},
}

@inproceedings{cumo,
  author       = {Jiachen Li and
                  Xinyao Wang and
                  Sijie Zhu and
                  Chia{-}Wen Kuo and
                  Lu Xu and
                  Fan Chen and
                  Jitesh Jain and
                  Humphrey Shi and
                  Longyin Wen},
  title        = {CuMo: Scaling Multimodal {LLM} with Co-Upcycled Mixture-of-Experts},
  booktitle    = {Advances in Neural Information Processing Systems},
  year         = {2024},
}

@article{tinyllava,
  title={Tinyllava: A framework of small-scale large multimodal models},
  author={Zhou, Baichuan and Hu, Ying and Weng, Xi and Jia, Junlong and Luo, Jie and Liu, Xien and Wu, Ji and Huang, Lei},
  journal={arXiv preprint arXiv:2402.14289},
  year={2024}
}

@article{moe-quant,
  author       = {Pingzhi Li and
                  Xiaolong Jin and
                  Yu Cheng and
                  Tianlong Chen},
  title        = {Examining Post-Training Quantization for Mixture-of-Experts: {A} Benchmark},
  journal      = {CoRR},
  volume       = {abs/2406.08155},
  year         = {2024},
}

@inproceedings{qmoe,
  author       = {Elias Frantar and
                  Dan Alistarh},
  title        = {QMoE: Sub-1-Bit Compression of Trillion Parameter Models},
  booktitle    = {Proceedings of the Seventh Annual Conference on Machine Learning and
                  Systems, MLSys 2024, Santa Clara, CA, USA, May 13-16, 2024},
  publisher    = {mlsys.org},
  year         = {2024},
}

@article{seedbench,
  author       = {Bohao Li and
                  Rui Wang and
                  Guangzhi Wang and
                  Yuying Ge and
                  Yixiao Ge and
                  Ying Shan},
  title        = {SEED-Bench: Benchmarking Multimodal LLMs with Generative Comprehension},
  journal      = {CoRR},
  volume       = {abs/2307.16125},
  year         = {2023},
}

@inproceedings{xnor,
  author       = {Mohammad Rastegari and
                  Vicente Ordonez and
                  Joseph Redmon and
                  Ali Farhadi},
  title        = {XNOR-Net: ImageNet Classification Using Binary Convolutional Neural
                  Networks},
  booktitle    = {Computer Vision - {ECCV} 2016 - 14th European Conference, Amsterdam,
                  The Netherlands, October 11-14, 2016, Proceedings, Part {IV}},
  series       = {Lecture Notes in Computer Science},
  volume       = {9908},
  pages        = {525--542},
  publisher    = {Springer},
  year         = {2016},
}

@inproceedings{binarybert,
  author       = {Haoli Bai and
                  Wei Zhang and
                  Lu Hou and
                  Lifeng Shang and
                  Jin Jin and
                  Xin Jiang and
                  Qun Liu and
                  Michael R. Lyu and
                  Irwin King},
  title        = {BinaryBERT: Pushing the Limit of {BERT} Quantization},
  booktitle    = {Proceedings of the 59th Annual Meeting of the Association for Computational
                  Linguistics and the 11th International Joint Conference on Natural
                  Language Processing, {ACL/IJCNLP} 2021, (Volume 1: Long Papers), Virtual
                  Event, August 1-6, 2021},
  pages        = {4334--4348},
  publisher    = {Association for Computational Linguistics},
  year         = {2021},
}
\bibliographystyle{alpha}

%%%%%%%%%%%%%%%%%%%%%%%%%%%%%%%%%%%%%%%%%%%%%%%%%%%%%%%%%%%%

\appendix

\section{Limitations}
\label{ap:lim}
In this work, we explore ternary MoE up-cycling for large multimodal models. Extensive experiments demonstrate that the proposed \our{} achieves both strong performance and significant memory savings compared to the widely adopted full-precision up-cycling baseline, MoE-LLaVA. However, this study does not provide a theoretical explanation for why ternary up-cycling can match the performance of its full-precision counterpart. We leave a deeper investigation into the training dynamics and theoretical underpinnings of ternary MoE models as future work.

\section{Hyper-parameters}
\label{ap:hyper}

In this section, we present the detailed hyper-parameters used for the training of \our{} and full-precision up-cycling baseline MoE-LLaVA. For Stage I and Stage II, we adopt the same training recipe, data and hyper-parameters, for both \our{} and MoE-LLaVA. For Stage III, we use the learning rate and scheduler recommended by MoE-LLaVA for full-precision training. For \our{}, following BitNet, we use a much large learning rate and two-stage weight decay for ternary experts which is critical for the optimization of extremely low-bit training. 

We utilize \textit{torch.compile} to compile the PyTorch code in the quantization into optimized kernels, which significantly speed up the training of \our{}. Above all, \our{} has similar training time compared to full-precision up-cycling MoE-LLaVA.

\begin{table*}[h]
    % \small
    \centering
    \caption{Hyper-parameters for the training of \our{} and MoE-LLaVA with 0.5B model. $a/b$ denotes the value of \our{}/MoE-LLaVA. $1+4$ denotes that the model has one shared expert and four routed experts.}
    \label{ap:tab:hy:0_5b}
    \begin{tabular}{l|ccc}
    \toprule
    \bf Hyper-parameter     &  \bf Stage I & \bf Stage II & \bf Stage III \\
    \midrule
    Learning rate     & 1e-3 & 5e-5 & 1.5e-4/5e-5 \\
    Batch Size & 256 & 128 & 256 \\
    Weight decay & \xmark & \xmark & 0.1$\rightarrow$ 0/\xmark\\
    Training steps & 2500 & 8000 & 12500 \\
    Training sequence & 1024 & 1024 & 2048 \\
    Vision sequence & \multicolumn{3}{c}{729} \\
    AdamW $\beta$ & \multicolumn{3}{c}{(0.9, 0.999)}\\
    AdamW $\epsilon$ & \multicolumn{3}{c}{1e-8} \\
    \midrule
    \# MoE layer & - & - & 24 \\
    \# Experts & - & - & 1+4 / 0+4\\
    \# Top-$k$ & - & - & 1+1 / 0+2\\
    \bottomrule
    \end{tabular}
\end{table*}

\begin{table*}[h]
    % \small
    \centering
    \caption{Hyper-parameters for the training of \our{} and MoE-LLaVA with 1.5B and 3B model. $a/b$ denotes the value of \our{}/MoE-LLaVA. $1+4$ denotes that the model has one shared expert and four routed experts.}
    \label{ap:tab:hy:1_5b}
    \begin{tabular}{l|ccc}
    \toprule
    \bf Hyper-parameter     &  \bf Stage I & \bf Stage II & \bf Stage III \\
    \midrule
    Learning rate     & 1e-3 & 2e-5 & 1e-4/2e-5 \\
    Batch Size & 256 & 128 & 256 \\
    Weight decay & \xmark & \xmark & 0.1$\rightarrow$ 0/\xmark\\
    Training steps & 2500 & 8000 & 12500 \\
    Training sequence & 1024 & 1024 & 2048 \\
    Vision sequence & \multicolumn{3}{c}{729} \\
    AdamW $\beta$ & \multicolumn{3}{c}{(0.9, 0.999)}\\
    AdamW $\epsilon$ & \multicolumn{3}{c}{1e-8} \\
    \midrule
    \# MoE layer & - & - & 28 \\
    \# Experts & - & - & 1+4 / 0+4\\
    \# Top-$k$ & - & - & 1+1 / 0+2\\
    \bottomrule
    \end{tabular}
\end{table*}

\section{More Ablation Studies}
\label{ap:abl}

We compare \our{} with the randomly initialized routed experts in Stage III. We evaluate the zero-shot performance of these models on a range of image understanding tasks, including MMMU, MMBench, AI2D, ChartQA, SeedBench-2-Plus and MMStar dataset. 

Table~\ref{ap:tab:bit_shard} shows the results of both methods in 0.5B, 1.5B and 3B model size. Initializing from FFN outperforms random initialization by a gain of 1.0\%, 1.5\% and 0.3\% average accuracy on end tasks in 0.5B, 1.5B and 3B model size, respectively. The results demonstrate that using the pre-trained full-precision FFN for \our{}'s initialization achieves better performance across various model size.

\begin{table*}[h]
    \centering
    % \small
    \caption{Ablations on the initialization methods of the routed experts for \our{} across different model sizes.}
    \label{ap:tab:bit_shard}
    % \scalebox{0.8}{
    \begin{tabular}{c|cccccccc}
    \toprule
    \makecell{\bf Initialize \\ \bf from FFN}  & \bf MMMU & \bf MMBench & \bf AI2D & \bf ChartQA & \bf SeedBench$^{2+}$ & \bf MMStar & \bf Avg. \\
    \midrule
    \multicolumn{6}{l}{\ \emph{0.5B Model Up-cycling}} \\
    \xmark & \bf 34.8 & 50.5 & \bf 55.2 & \bf 55.8 & 43.0 & \bf 39.1 & 46.4 \\
    \rowcolor{myblue}\cmark & 34.2 & \bf 57.6 & \bf 55.2 & 54.9 & \bf 44.8 & 37.9 & \bf 47.4 \\
    \midrule
    \multicolumn{6}{l}{\ \emph{1.5B Model Up-cycling}} \\
    \xmark & 40.1 & 69.9 & 67.1 & 59.9 & 53.2 & 44.5 & 55.8 \\
    \rowcolor{myblue}\cmark & \bf 42.6 & \bf 70.0 & \bf 68.7 & \bf 61.3 & \bf 54.8 & \bf 46.4 & \bf 57.3 \\
    \midrule
    \multicolumn{6}{l}{\ \emph{3B Model Up-cycling}} \\
    \xmark & 43.3 & \bf 75.5 & 72.7 & 65.5 & 57.1 & \bf 48.8 & 60.5 \\
    \rowcolor{myblue}\cmark & \bf 43.4 & 74.5 & \bf 73.9 & \bf 67.6 & \bf 57.5 & 48.2 & \bf 60.8 \\
    \bottomrule
    \end{tabular}
    % }
\end{table*}

\section{Visualization}
\label{ap:vis}

We visualize the workflows of \our{}-1.5B at three distinct levels: expert, modality, and token. Specifically, we selected the AI2D, SeedBench-2-Plus, ChartQA, DocVQA, InfoVQA, MMStar, and MMBench datasets. Figures~\ref{ap:vis:fig1}, \ref{ap:vis:fig2}, and \ref{ap:vis:fig3} respectively illustrate the load distributions across different experts, the modality-aware routing distributions for each expert, and the top-10 activated pathways obtained via PCA. Our analysis indicates that, although the routing distributions of \our{} remain quite similar across tasks, they are predominantly influenced by the input modality.

\subsection{Routing distribution for tokens}
\label{ap:vis:rt}

\begin{figure*}[h]
    \centering
    \begin{subfigure}{0.325\textwidth}
        \centering
        \includegraphics[width=\textwidth]{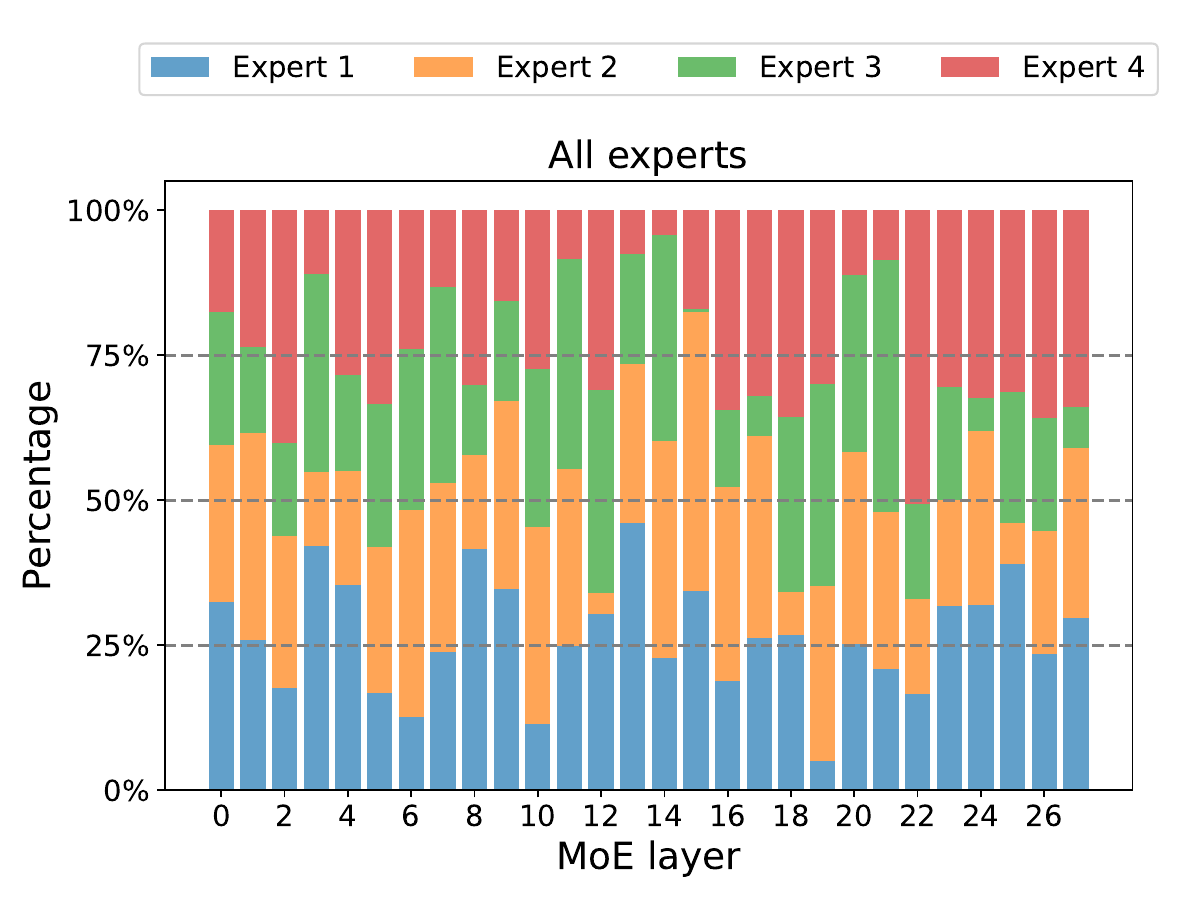}
        \caption{All tokens (AI2D)}
    \end{subfigure}
    \begin{subfigure}{0.325\textwidth}
        \centering
        \includegraphics[width=\textwidth]{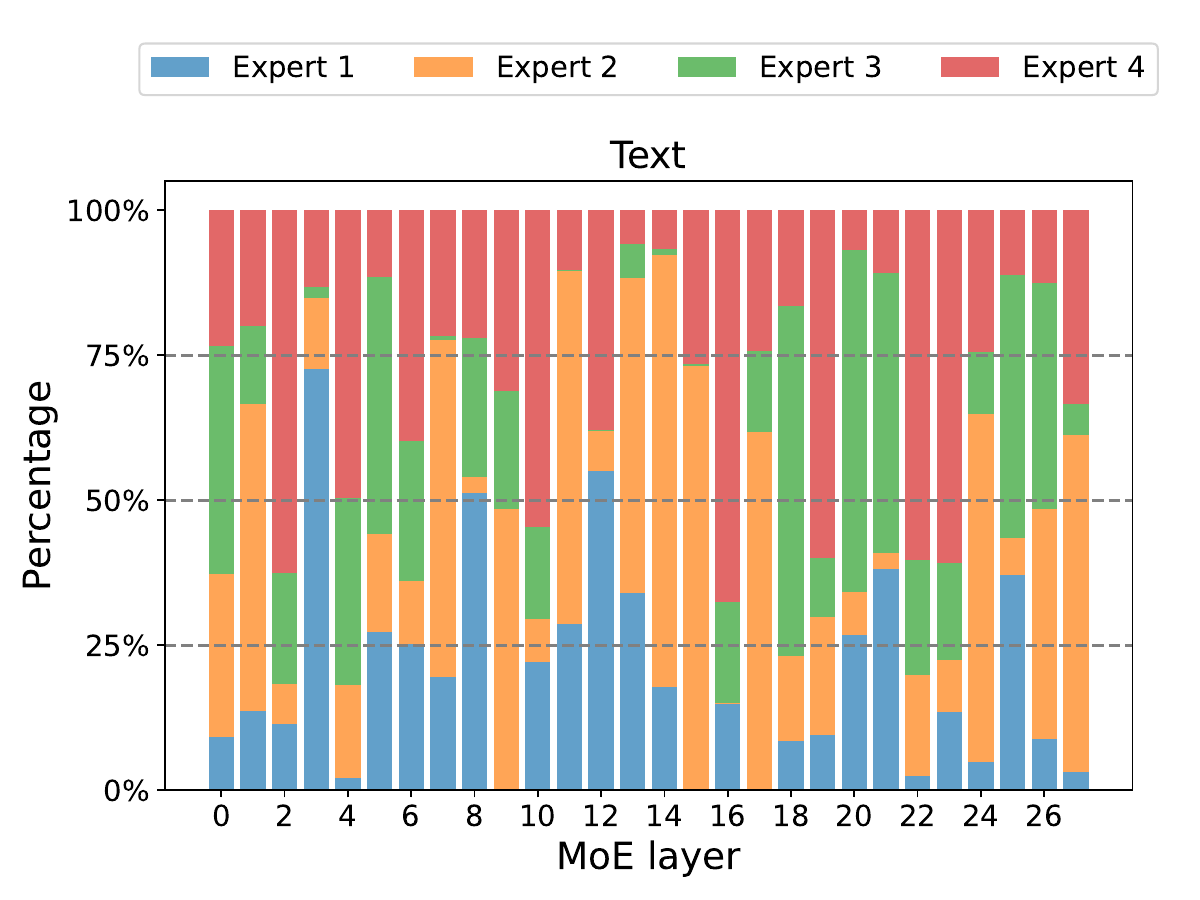}
        \caption{Text tokens (AI2D)}
    \end{subfigure}
    \begin{subfigure}{0.325\textwidth}
        \centering
        \includegraphics[width=\textwidth]{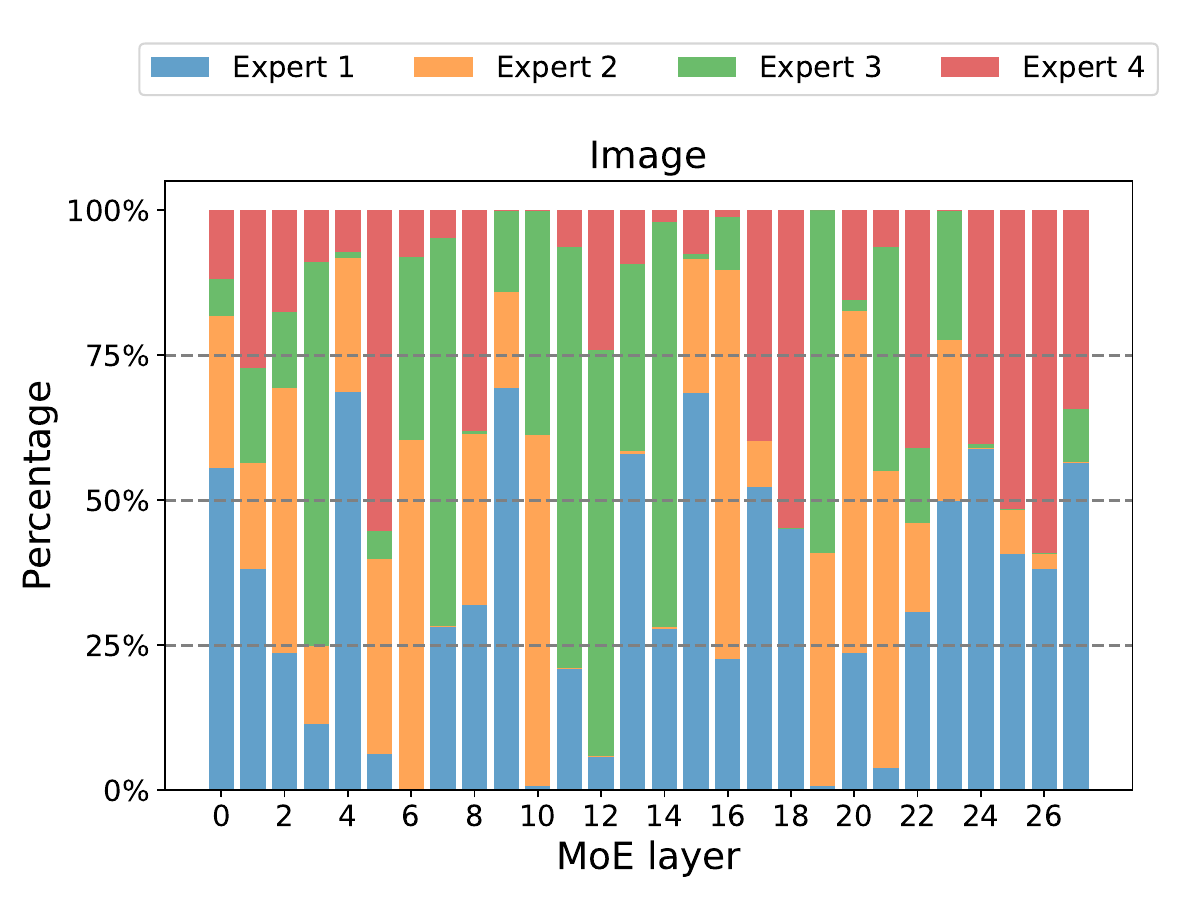}
        \caption{Image tokens (AI2D)}
    \end{subfigure}

    \begin{subfigure}{0.325\textwidth}
        \centering
        \includegraphics[width=\textwidth]{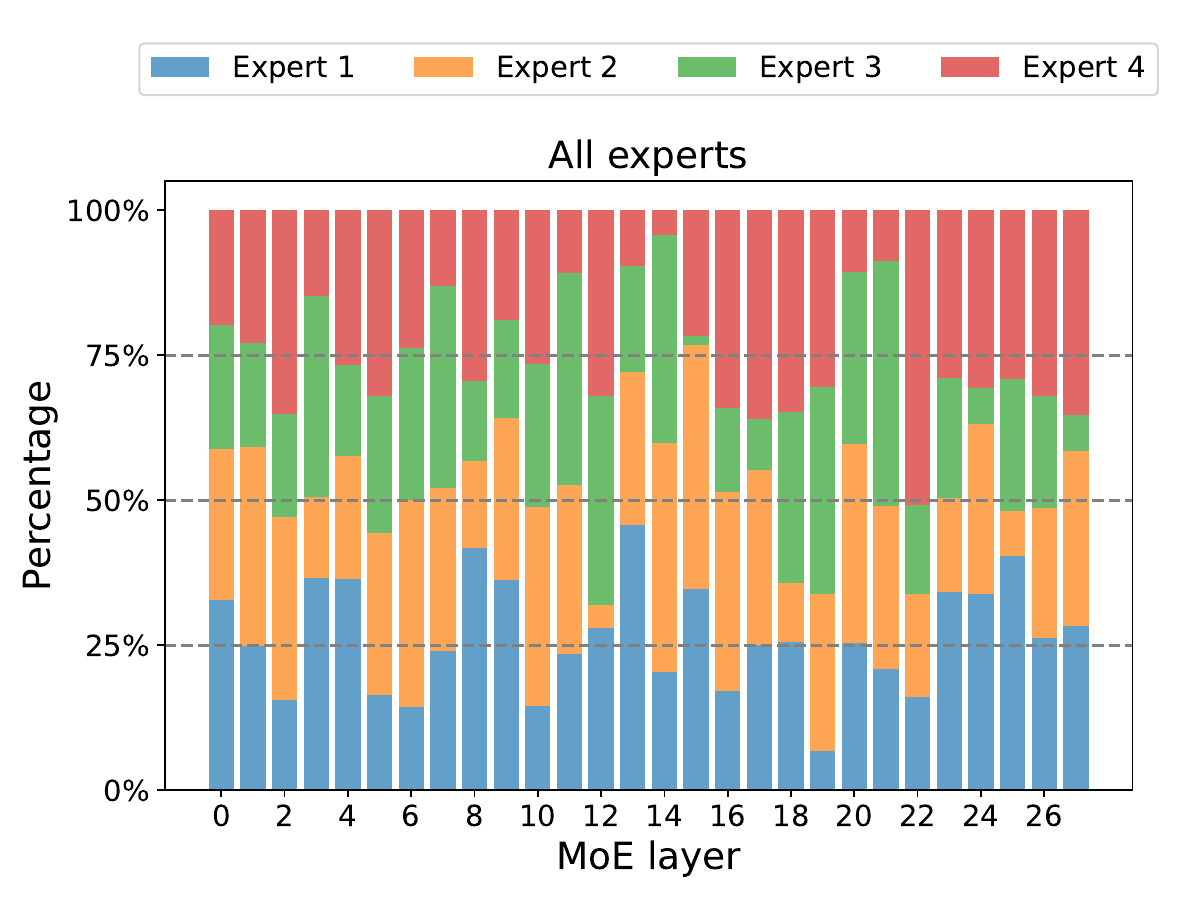}
        \caption{All tokens (SeedBench$^{2+}$)}
    \end{subfigure}
    \begin{subfigure}{0.325\textwidth}
        \centering
        \includegraphics[width=\textwidth]{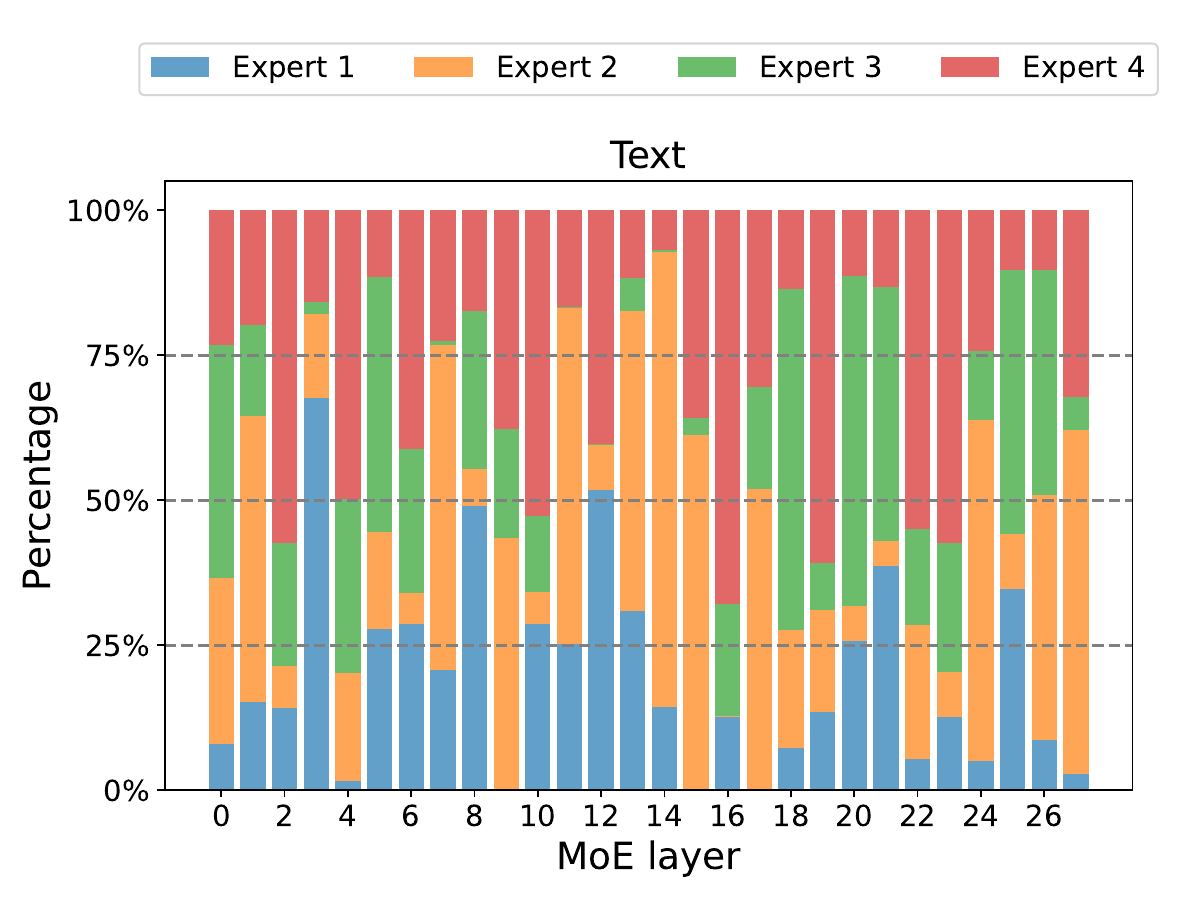}
        \caption{Text tokens (SeedBench$^{2+}$)}
    \end{subfigure}
    \begin{subfigure}{0.325\textwidth}
        \centering
        \includegraphics[width=\textwidth]{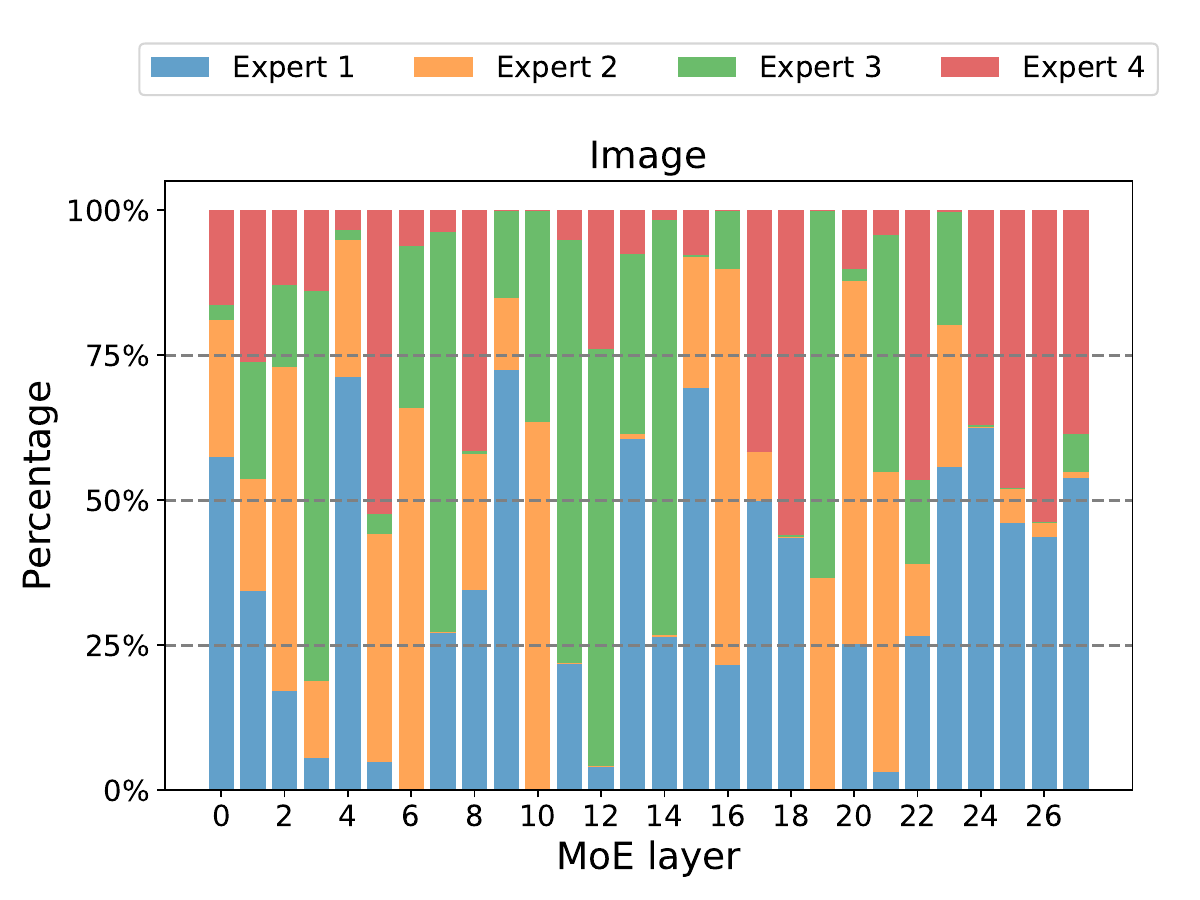}
        \caption{Image tokens (SeedBench$^{2+}$)}
    \end{subfigure}

    \begin{subfigure}{0.325\textwidth}
        \centering
        \includegraphics[width=\textwidth]{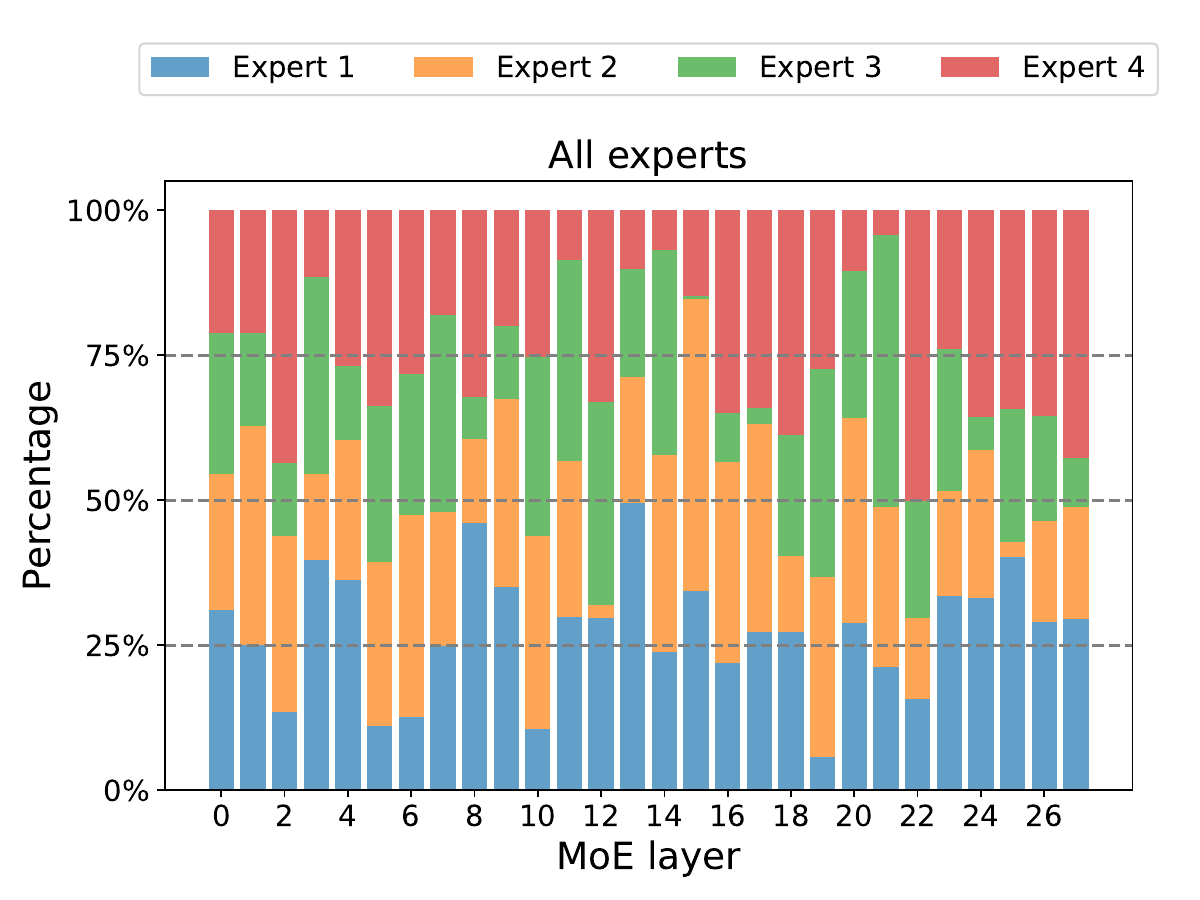}
        \caption{All tokens (ChartQA)}
    \end{subfigure}
    \begin{subfigure}{0.325\textwidth}
        \centering
        \includegraphics[width=\textwidth]{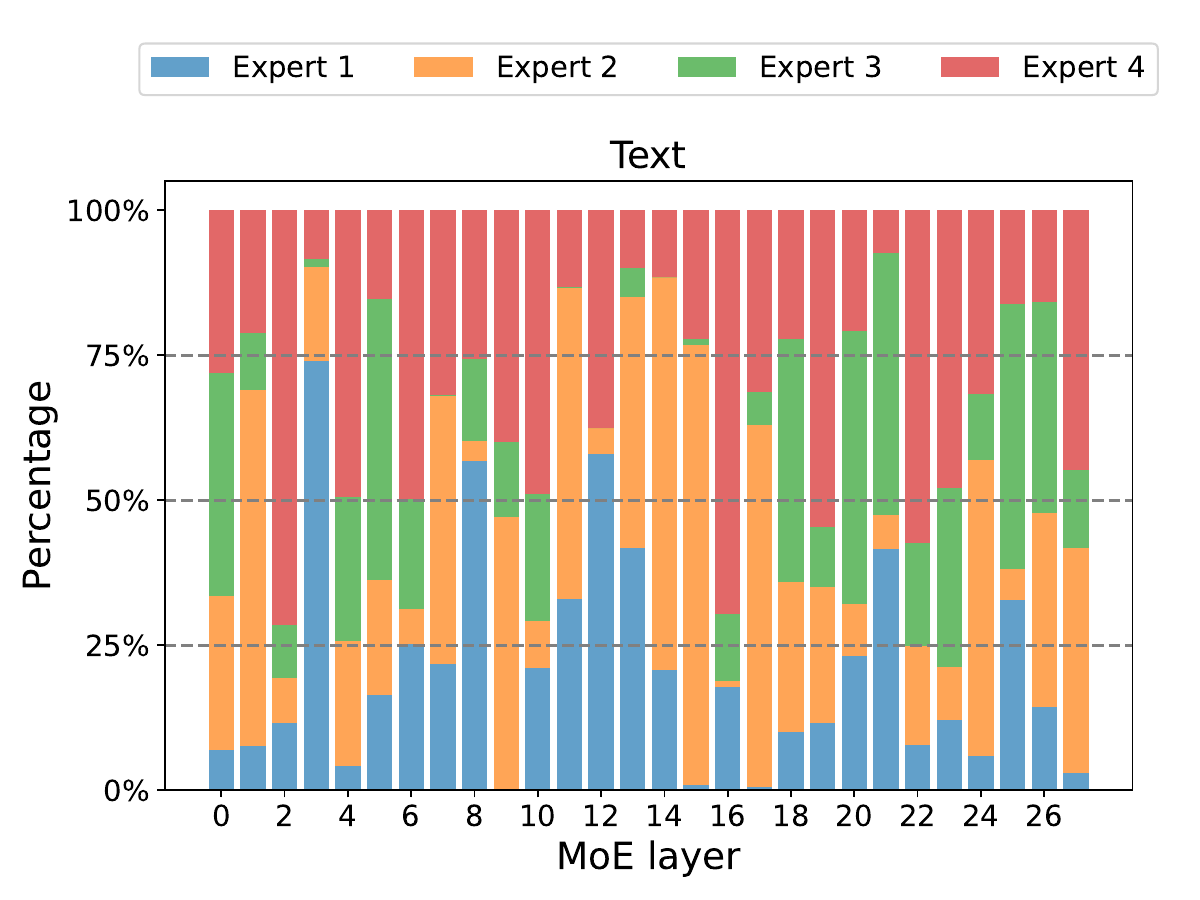}
        \caption{Text tokens (ChartQA)}
    \end{subfigure}
    \begin{subfigure}{0.325\textwidth}
        \centering
        \includegraphics[width=\textwidth]{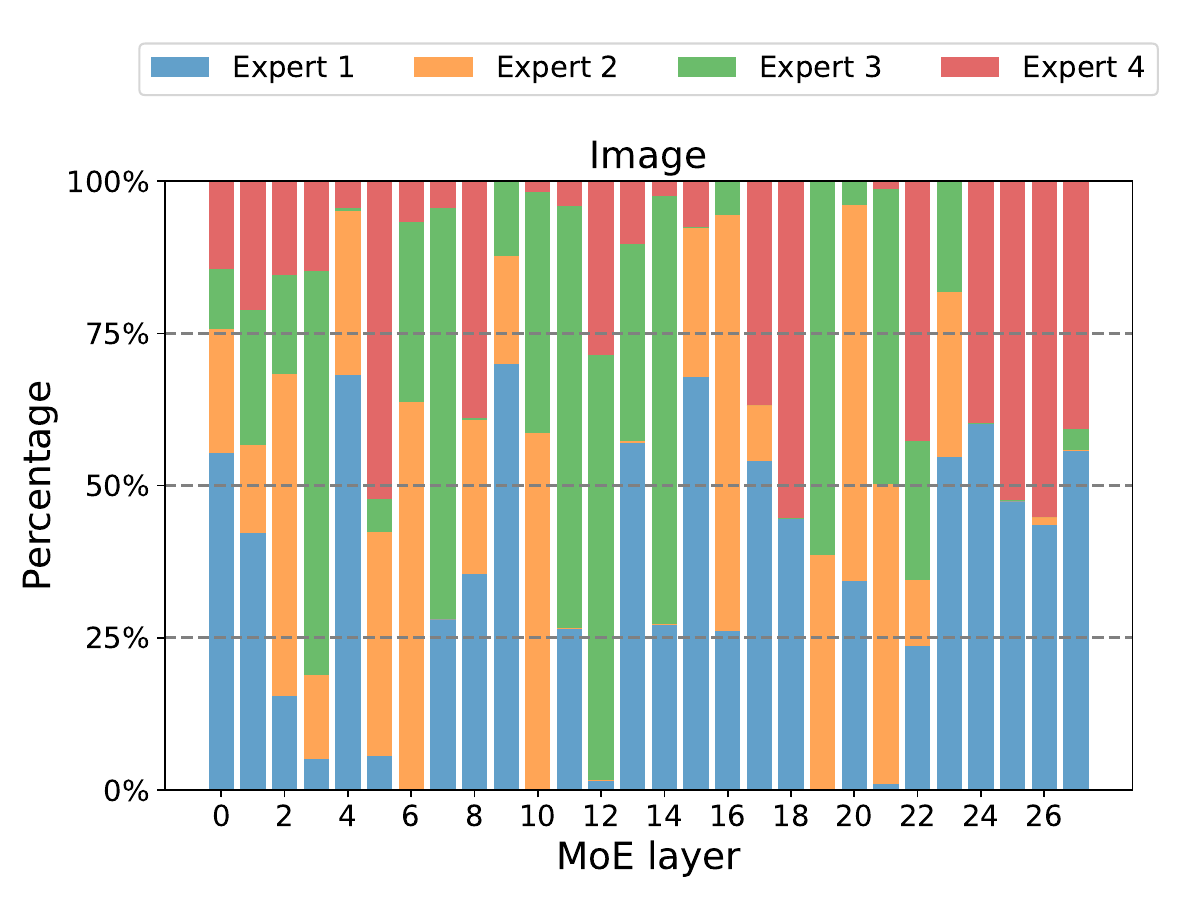}
        \caption{Image tokens (ChartQA)}
    \end{subfigure}

\end{figure*}

\begin{figure*}[h]
    \ContinuedFloat
    \centering
    \begin{subfigure}{0.325\textwidth}
        \centering
        \includegraphics[width=\textwidth]{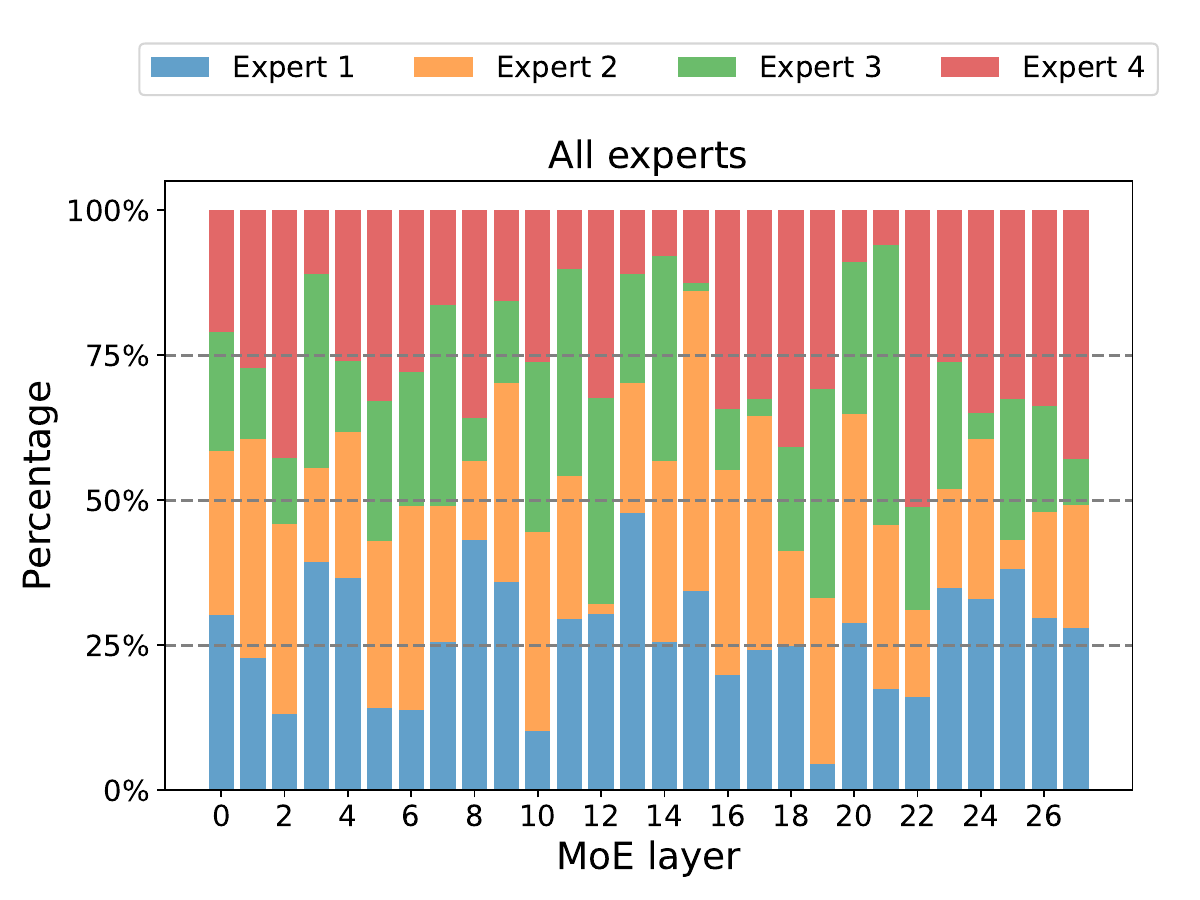}
        \caption{All tokens (DocVQA)}
    \end{subfigure}
    \begin{subfigure}{0.325\textwidth}
        \centering
        \includegraphics[width=\textwidth]{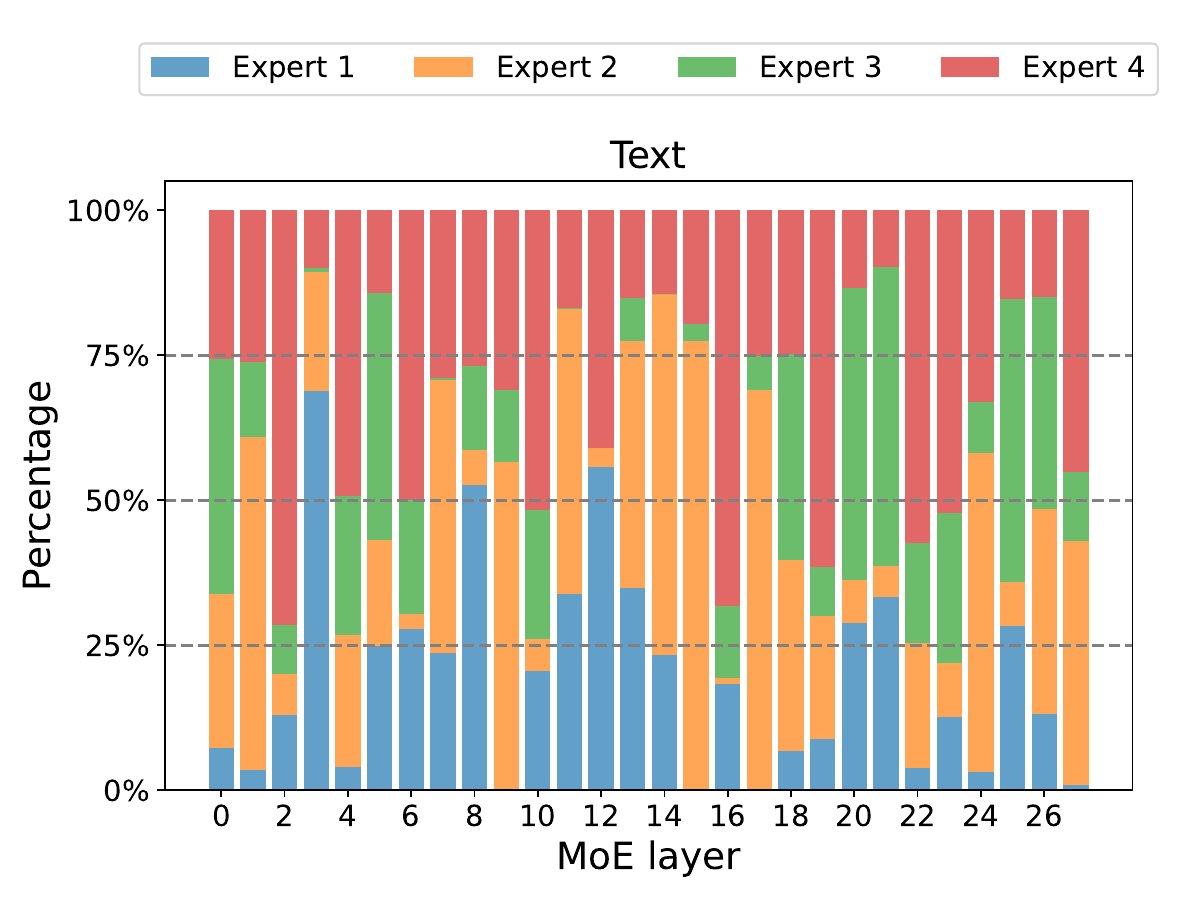}
        \caption{Text tokens (DocVQA)}
    \end{subfigure}
    \begin{subfigure}{0.325\textwidth}
        \centering
        \includegraphics[width=\textwidth]{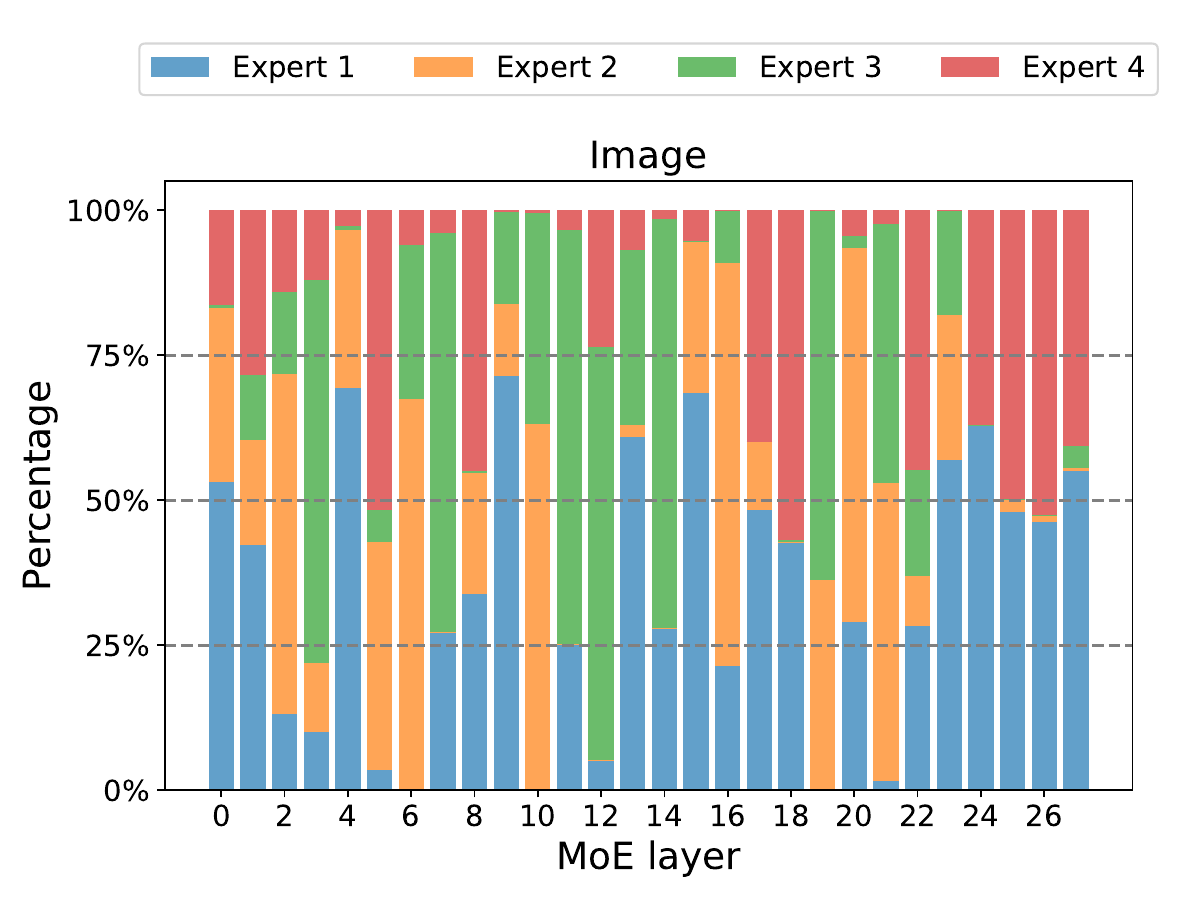}
        \caption{Image tokens (DocVQA)}
    \end{subfigure}
    
    \begin{subfigure}{0.325\textwidth}
        \centering
        \includegraphics[width=\textwidth]{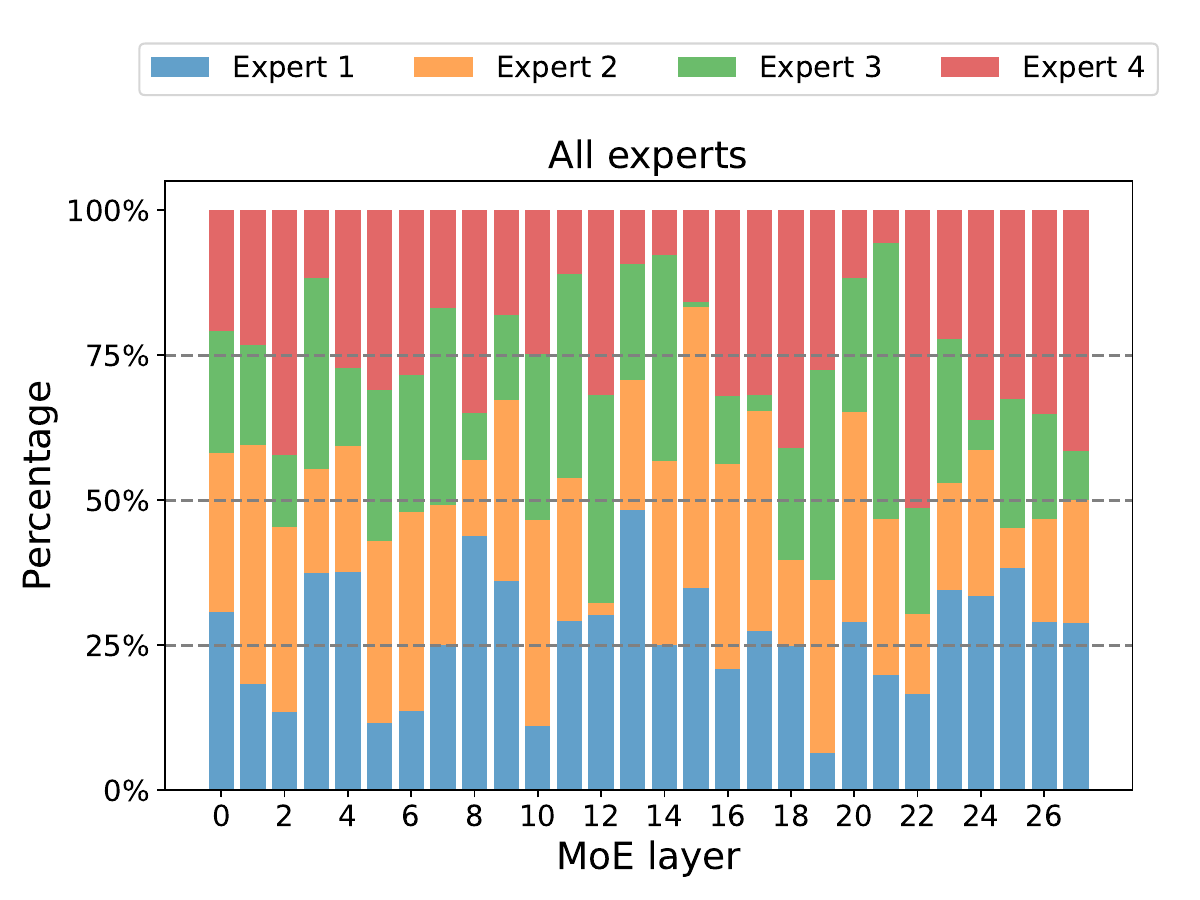}
        \caption{All tokens (InfoVQA)}
    \end{subfigure}
    \begin{subfigure}{0.325\textwidth}
        \centering
        \includegraphics[width=\textwidth]{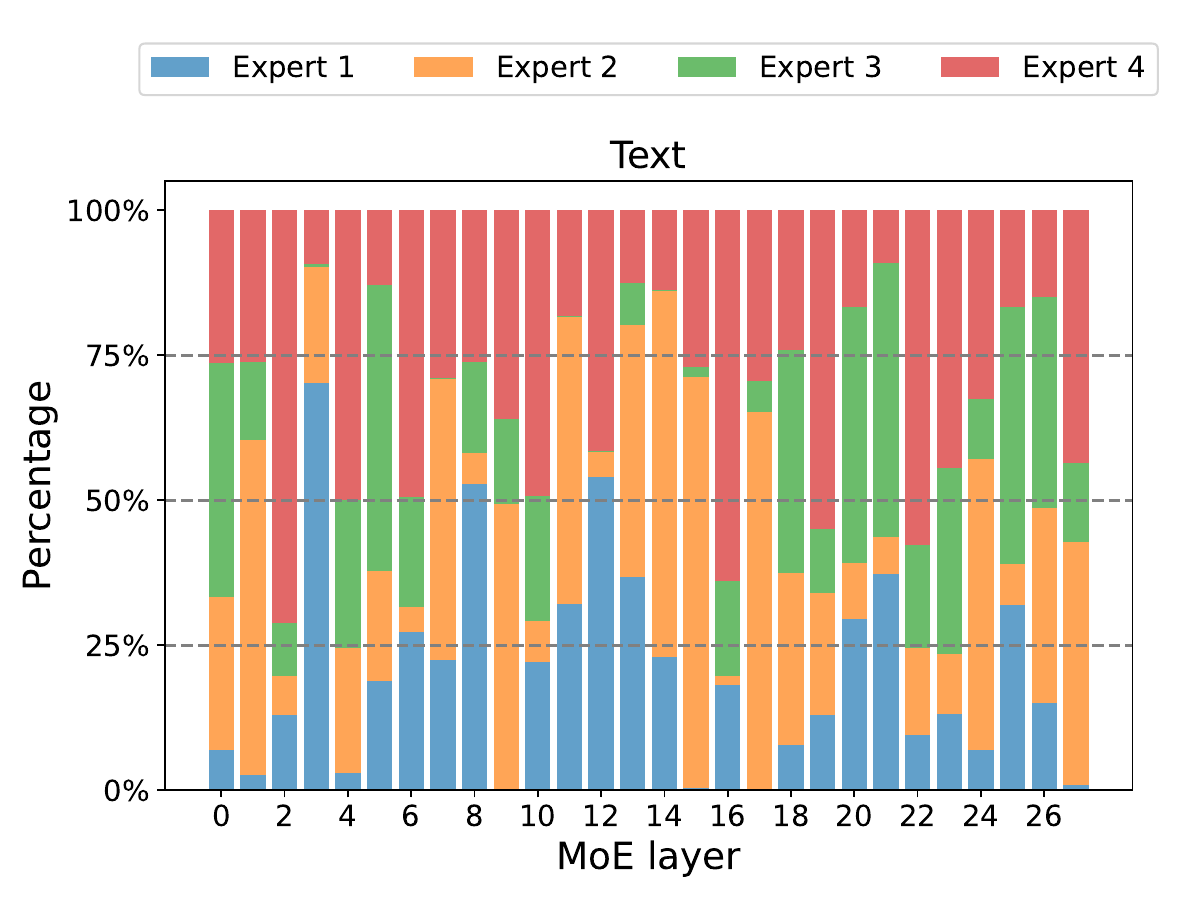}
        \caption{Text tokens (InfoVQA)}
    \end{subfigure}
    \begin{subfigure}{0.325\textwidth}
        \centering
        \includegraphics[width=\textwidth]{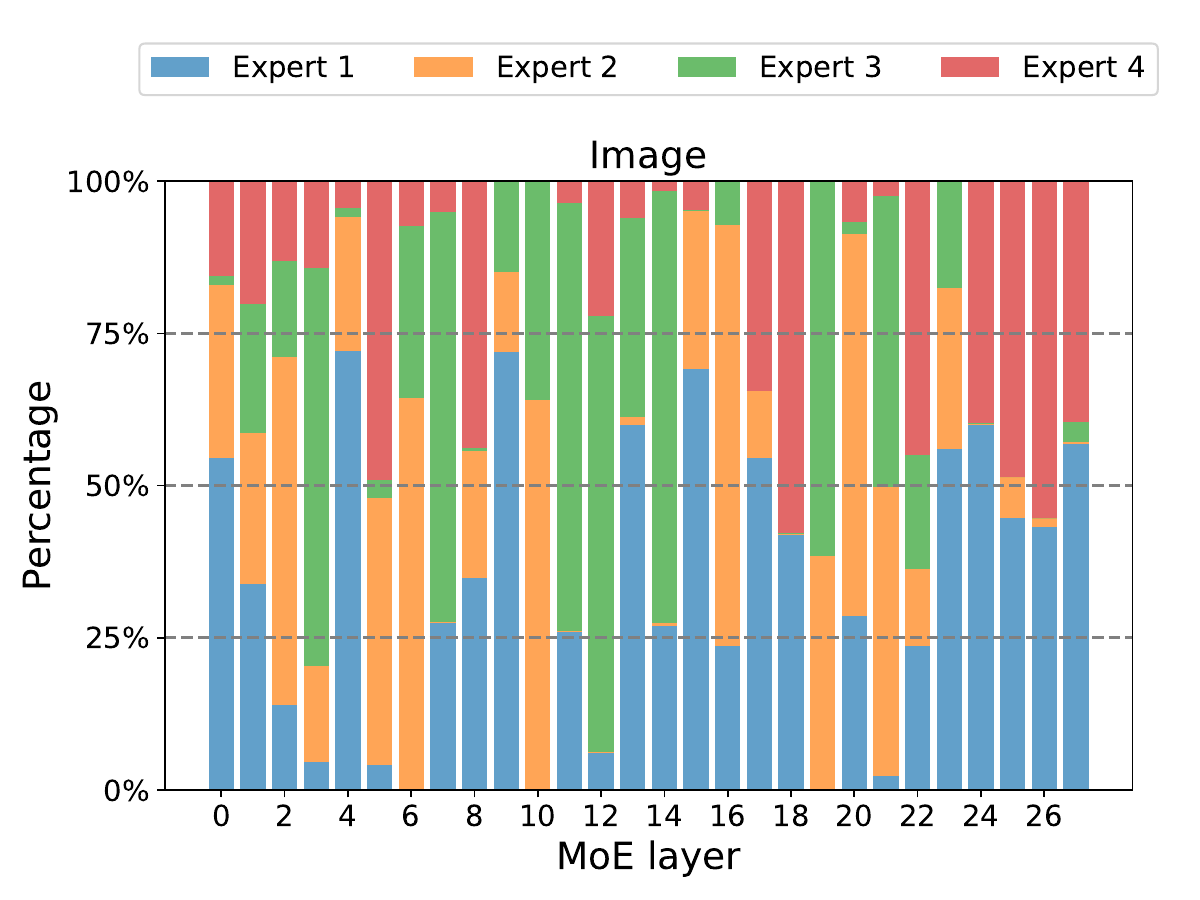}
        \caption{Image tokens (InfoVQA)}
    \end{subfigure}
    \begin{subfigure}{0.325\textwidth}
        \centering
        \includegraphics[width=\textwidth]{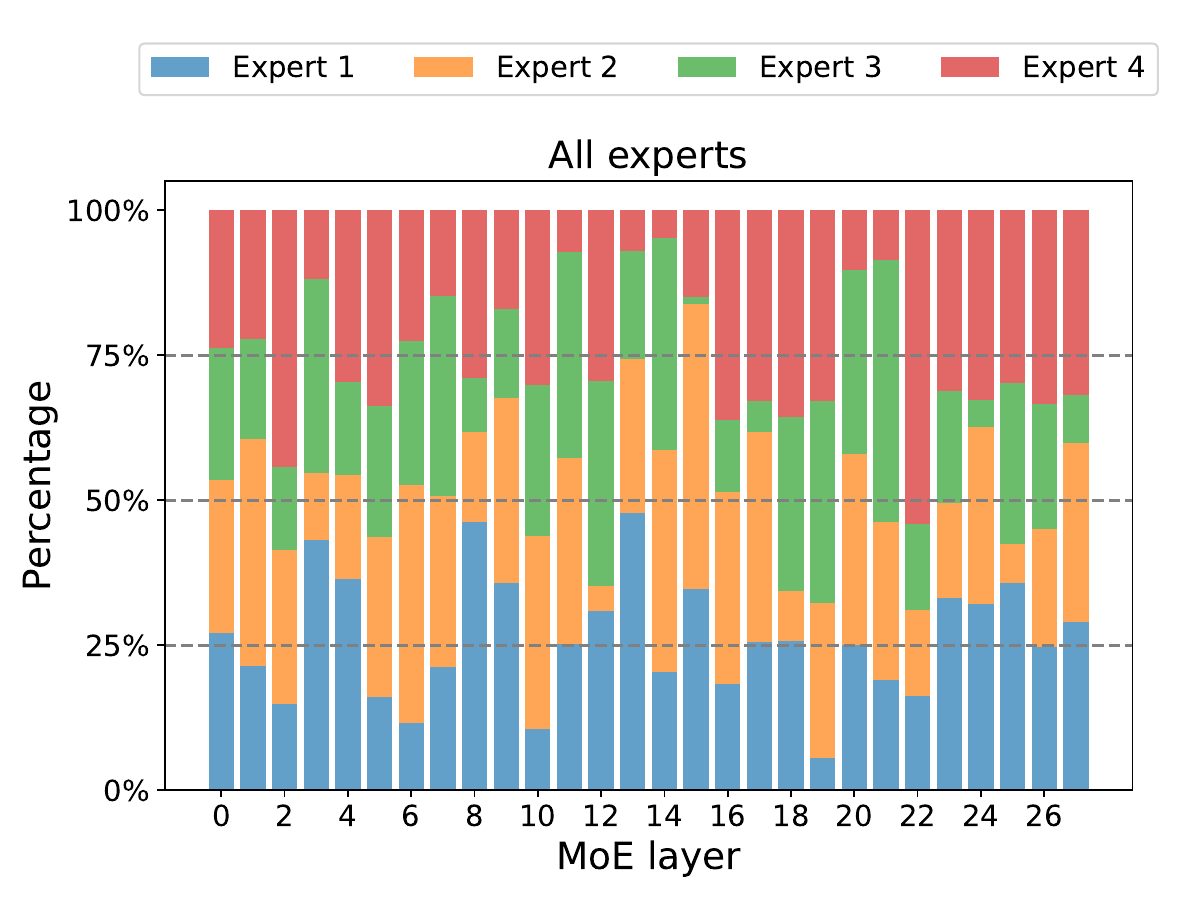}
        \caption{All tokens (MMStar)}
    \end{subfigure}
    \begin{subfigure}{0.325\textwidth}
        \centering
        \includegraphics[width=\textwidth]{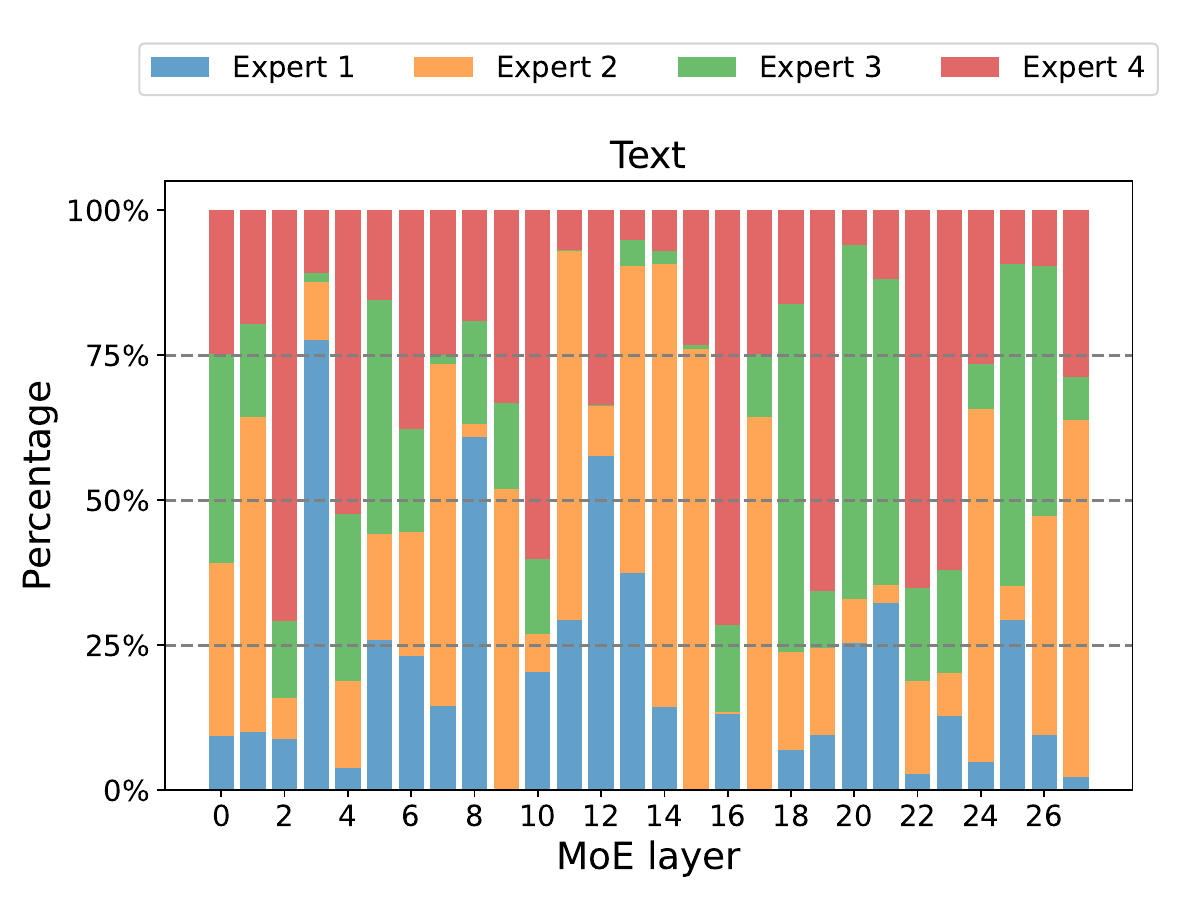}
        \caption{Text tokens (MMStar)}
    \end{subfigure}
    \begin{subfigure}{0.325\textwidth}
        \centering
        \includegraphics[width=\textwidth]{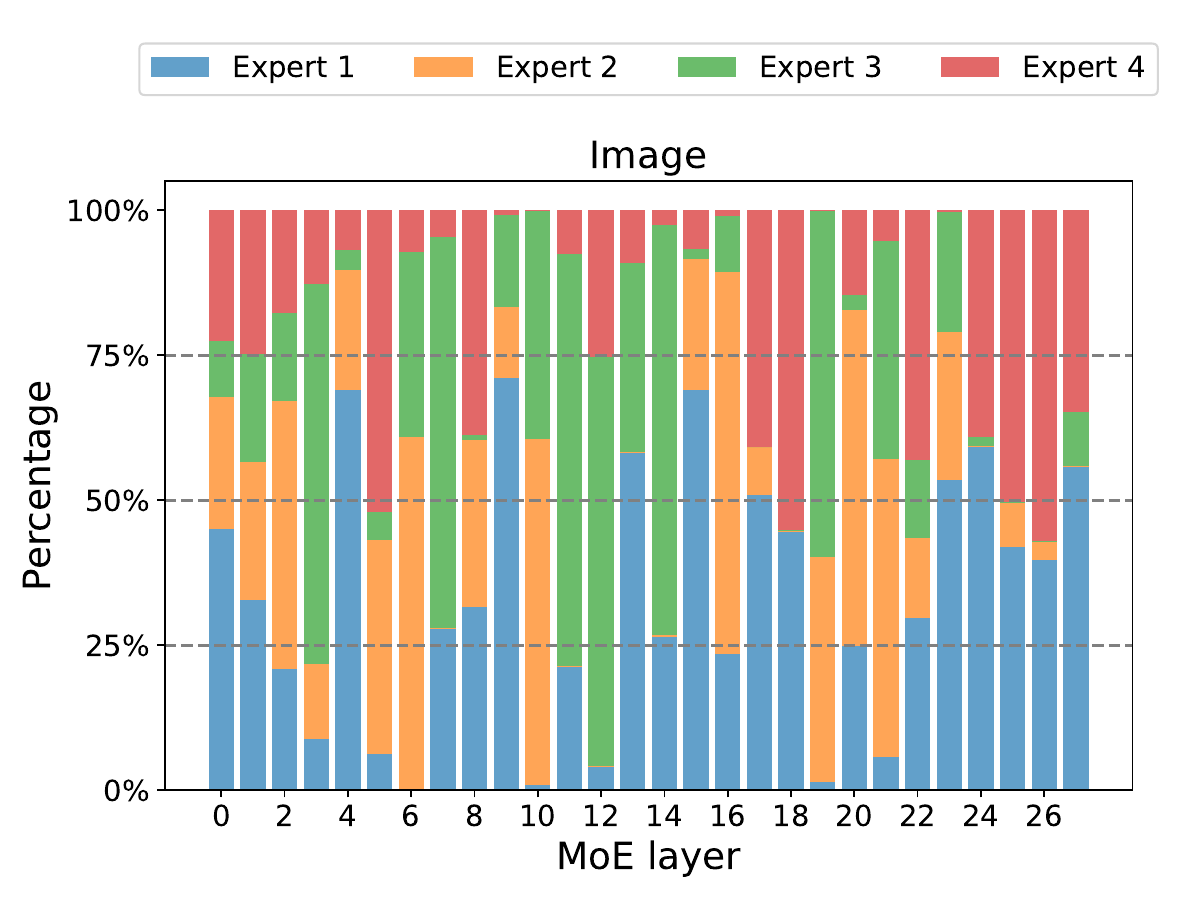}
        \caption{Image tokens (MMStar)}
    \end{subfigure}
    
    \begin{subfigure}{0.325\textwidth}
        \centering
        \includegraphics[width=\textwidth]{1.5b-mote-10m/mmbench_en_test-fig2_1.pdf}
        \caption{All tokens (MMBench)}
    \end{subfigure}
    \begin{subfigure}{0.325\textwidth}
        \centering
        \includegraphics[width=\textwidth]{1.5b-mote-10m/mmbench_en_test-fig2_2.pdf}
        \caption{Text tokens (MMBench)}
    \end{subfigure}
    \begin{subfigure}{0.325\textwidth}
        \centering
        \includegraphics[width=\textwidth]{1.5b-mote-10m/mmbench_en_test-fig2_3.pdf}
        \caption{Image tokens (MMBench)}
    \end{subfigure}
    \caption{Visualization of the routing distributions of all tokens, text tokens, image tokens across all experts on various tasks.}
    \label{ap:vis:fig1}
\end{figure*}

\clearpage

\subsection{Routing distribution for each experts}
\label{ap:vis:expert}

\begin{figure*}[h]
    \centering
    \begin{subfigure}{\textwidth}
        \centering
        \includegraphics[width=\textwidth]{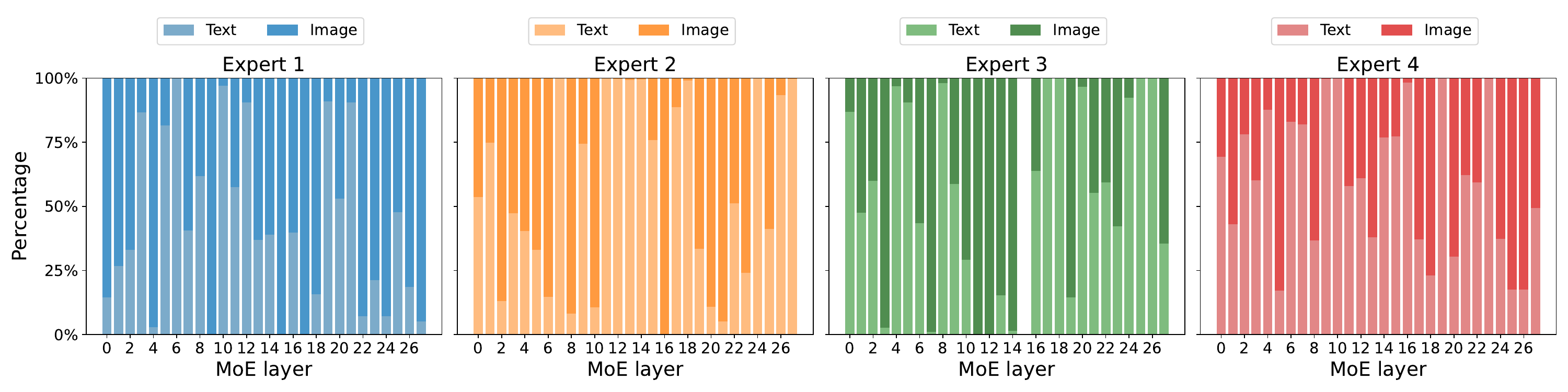}
        \caption{Routing distribution on AI2D.}
    \end{subfigure}

    \begin{subfigure}{\textwidth}
        \centering
        \includegraphics[width=\textwidth]{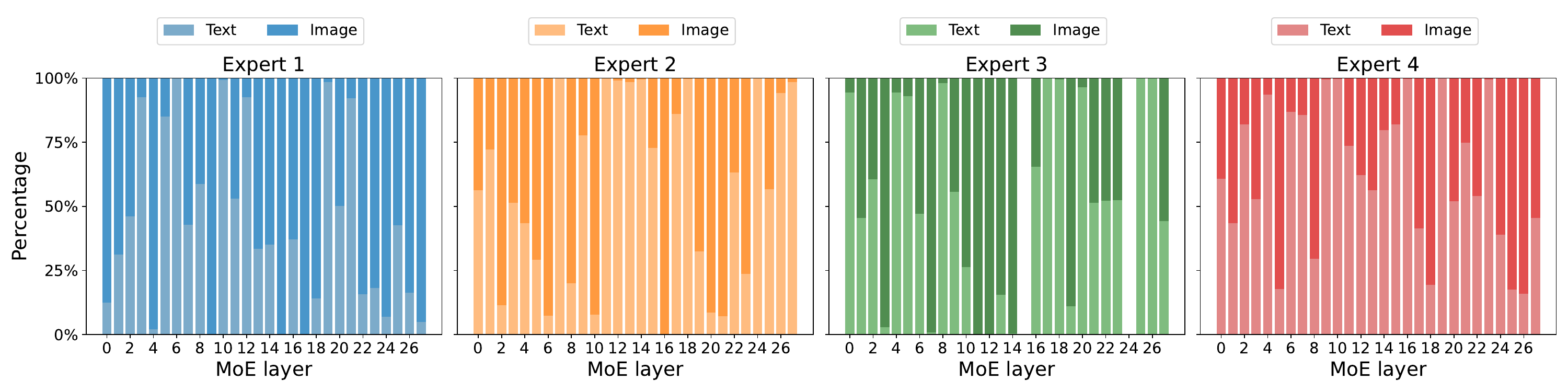}
        \caption{Routing distribution on SeedBench-2-Plus.}
    \end{subfigure}
    
    \begin{subfigure}{\textwidth}
        \centering
        \includegraphics[width=\textwidth]{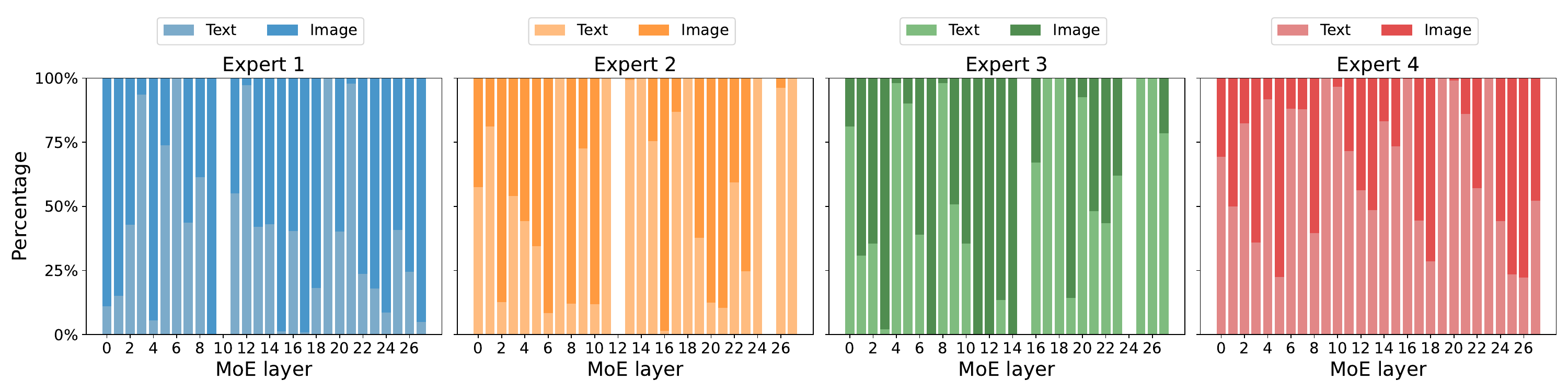}
        \caption{Routing distribution on ChartQA.}
    \end{subfigure}
    
    \begin{subfigure}{\textwidth}
        \centering
        \includegraphics[width=\textwidth]{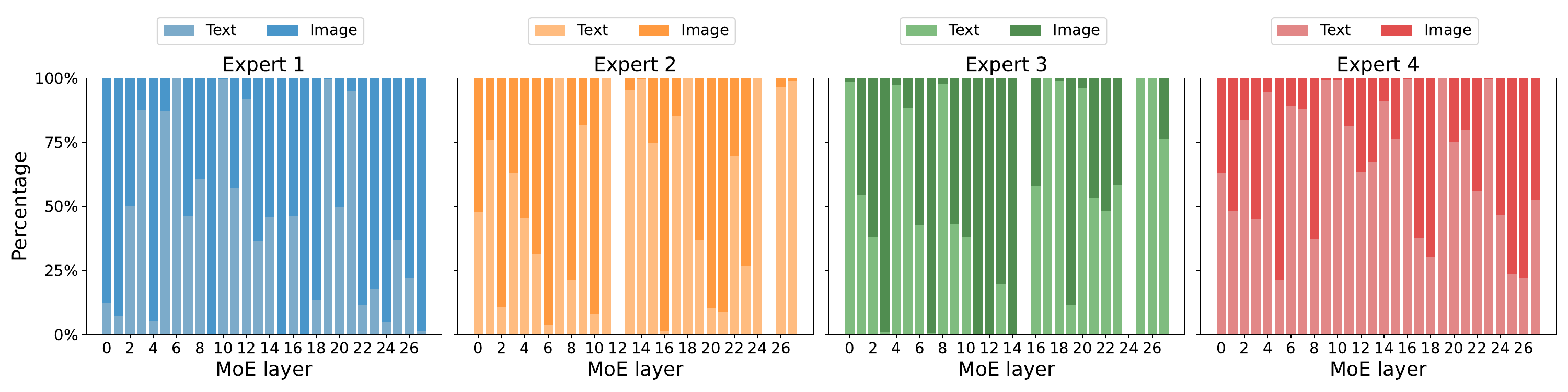}
        \caption{Routing distribution on DocVQA.}
    \end{subfigure}

\end{figure*}

\begin{figure*}[h]
    \ContinuedFloat
    \centering
    \begin{subfigure}{\textwidth}
        \centering
        \includegraphics[width=\textwidth]{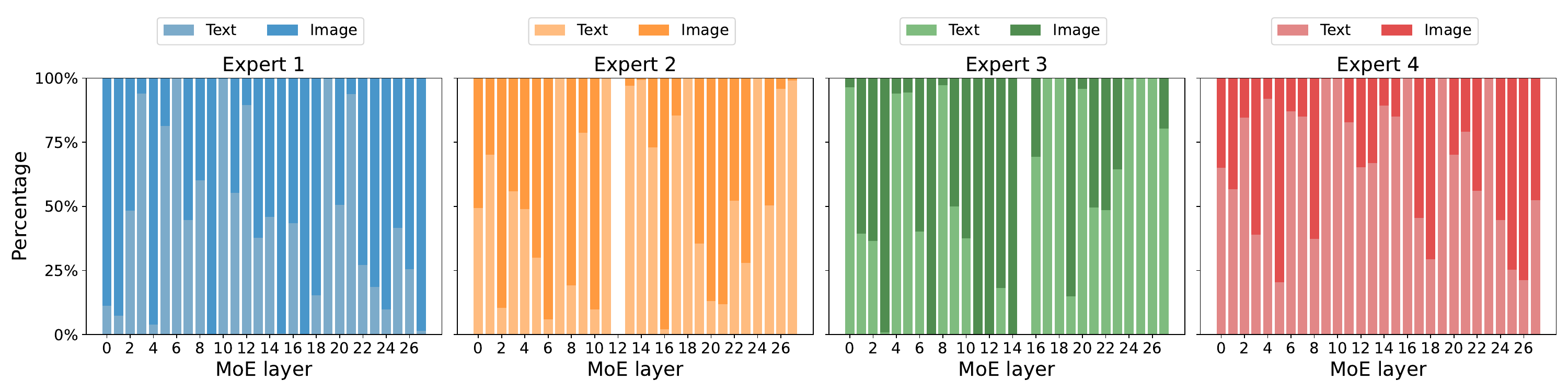}
        \caption{Routing distribution on InfoVQA.}
    \end{subfigure}
    
    \begin{subfigure}{\textwidth}
        \centering
        \includegraphics[width=\textwidth]{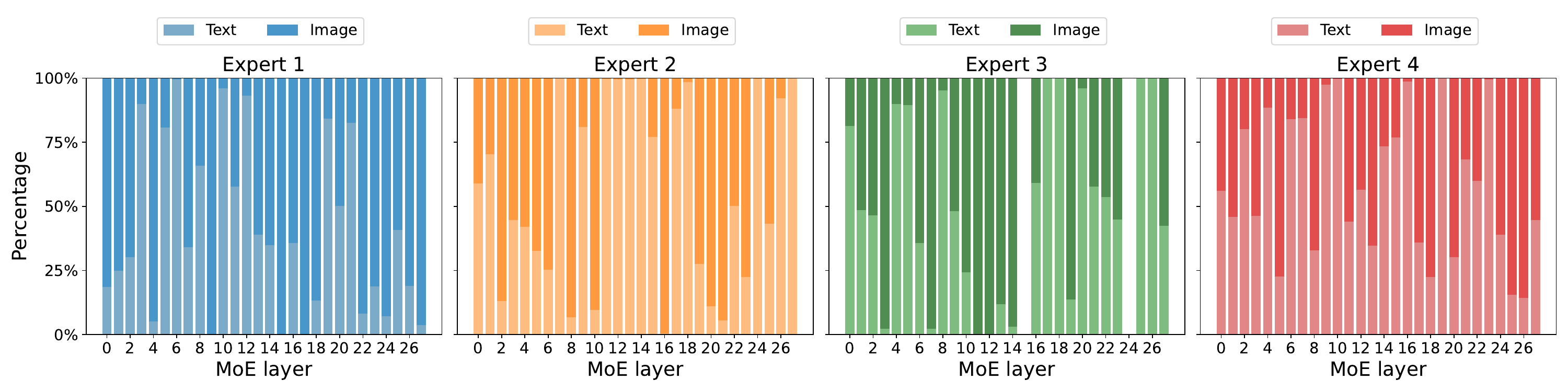}
        \caption{Routing distribution on MMStar.}
    \end{subfigure}
    
    \begin{subfigure}{\textwidth}
        \centering
        \includegraphics[width=\textwidth]{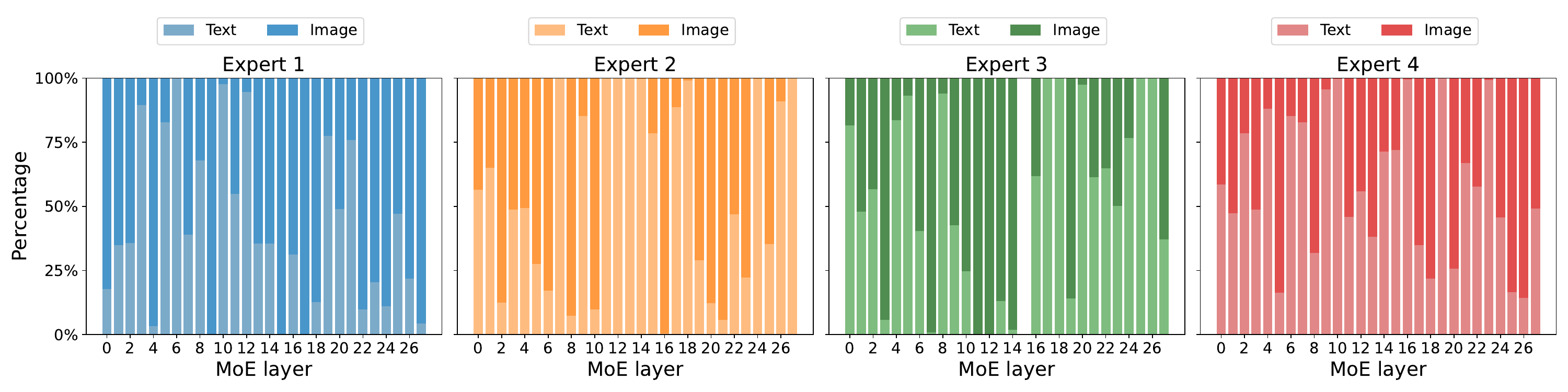}
        \caption{Routing distribution on MMBench.}
    \end{subfigure}
    \caption{Visualization of the modality-aware routing distributions for each expert on various tasks.}
    \label{ap:vis:fig2}
\end{figure*}

\clearpage

\subsection{Activated Pathways}
\label{ap:vis:path}

\begin{figure*}[h]
    \centering
    \begin{subfigure}{0.8\textwidth}
        \centering
        \includegraphics[width=\textwidth]{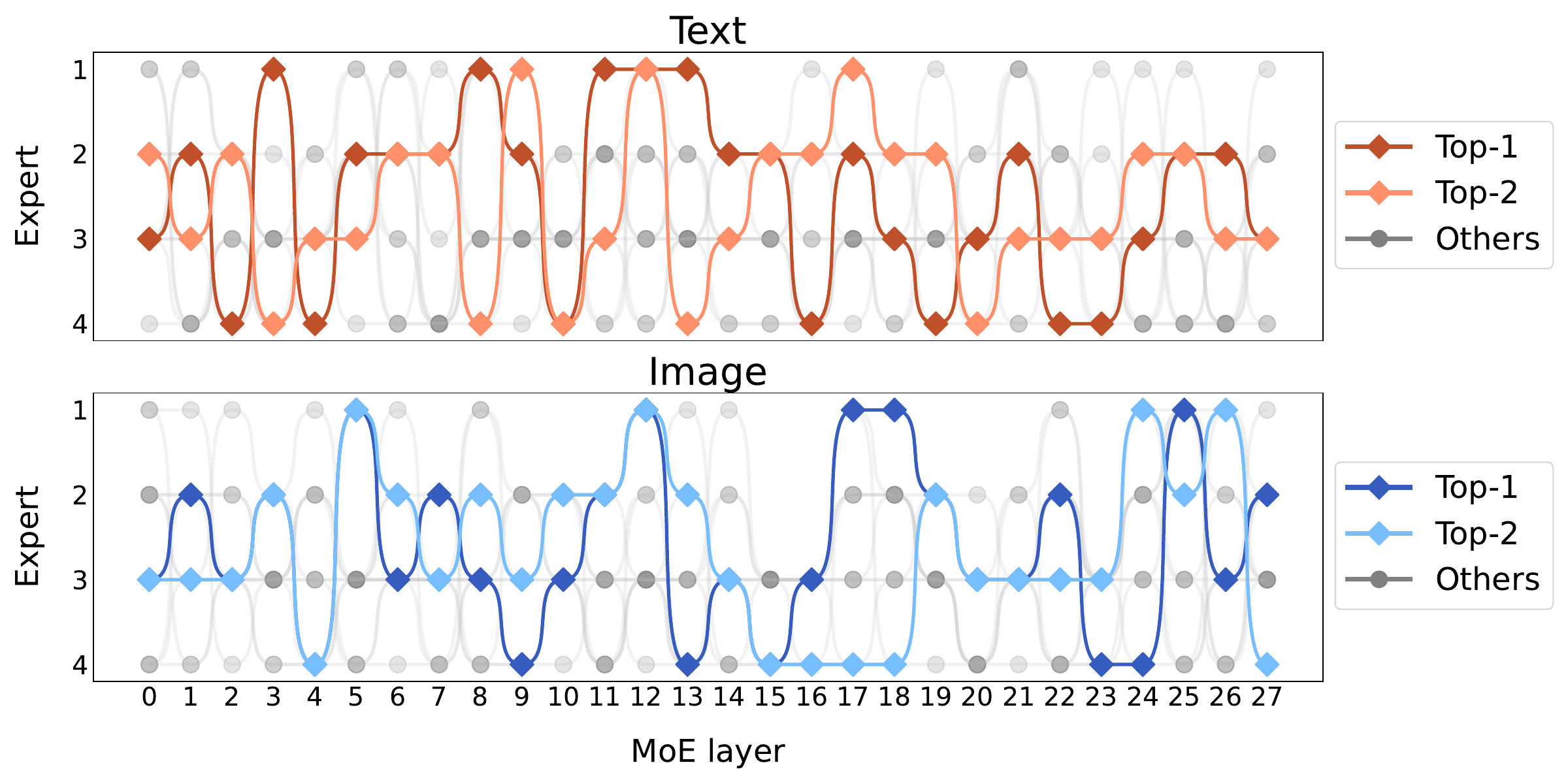}
        \caption{The top-10 pathways for text and image tokens on MMBench.}
    \end{subfigure}
    
    \begin{subfigure}{0.8\textwidth}
        \centering
        \includegraphics[width=\textwidth]{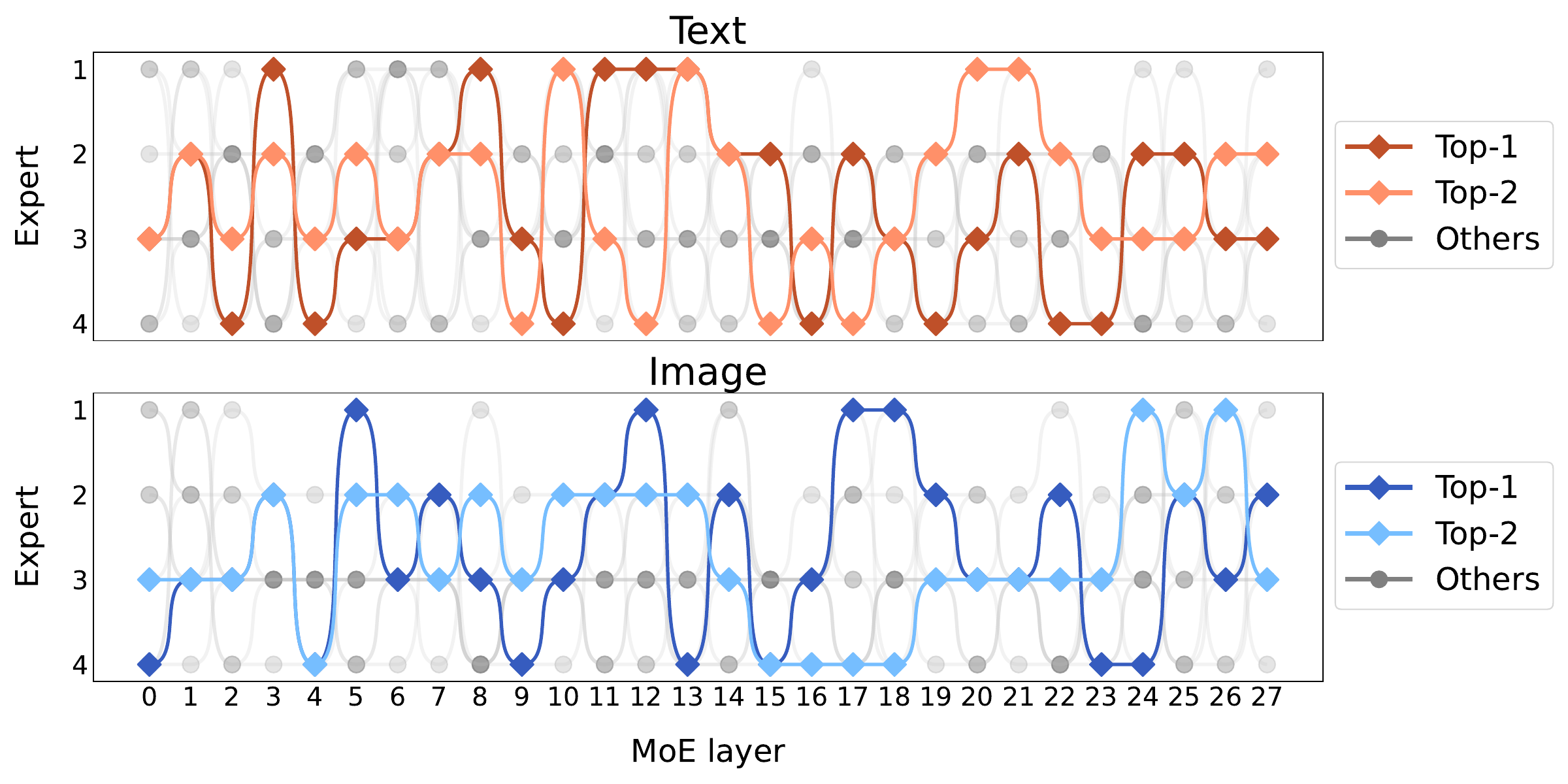}
        \caption{The top-10 pathways for text and image tokens on AI2D.}
    \end{subfigure}

    \begin{subfigure}{0.8\textwidth}
        \centering
        \includegraphics[width=\textwidth]{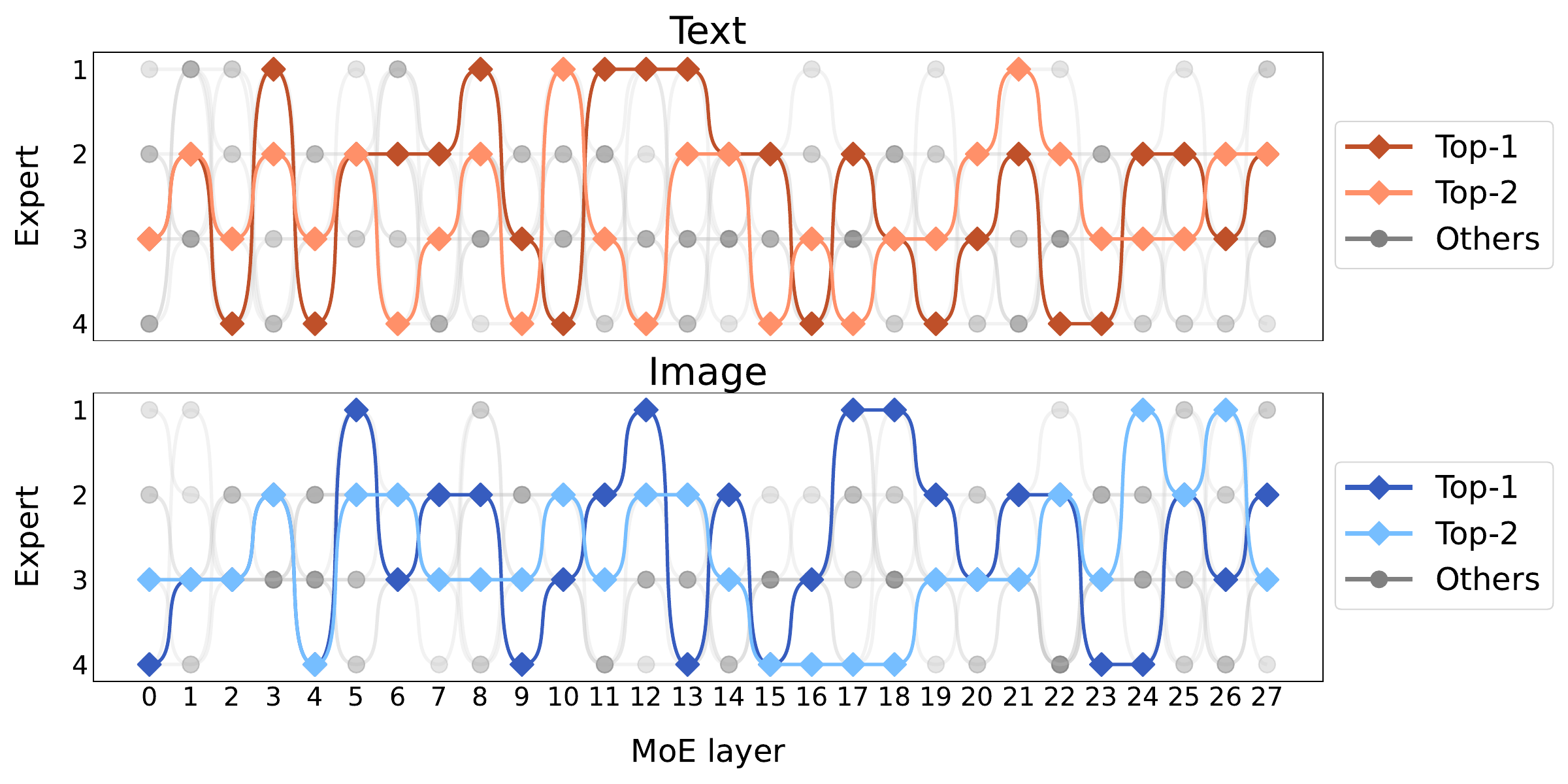}
        \caption{The top-10 pathways for text and image tokens on SeedBench-2-Plus.}
    \end{subfigure}
    
\end{figure*}

\begin{figure*}[h]
    \ContinuedFloat
    \centering
    \begin{subfigure}{0.8\textwidth}
        \centering
        \includegraphics[width=\textwidth]{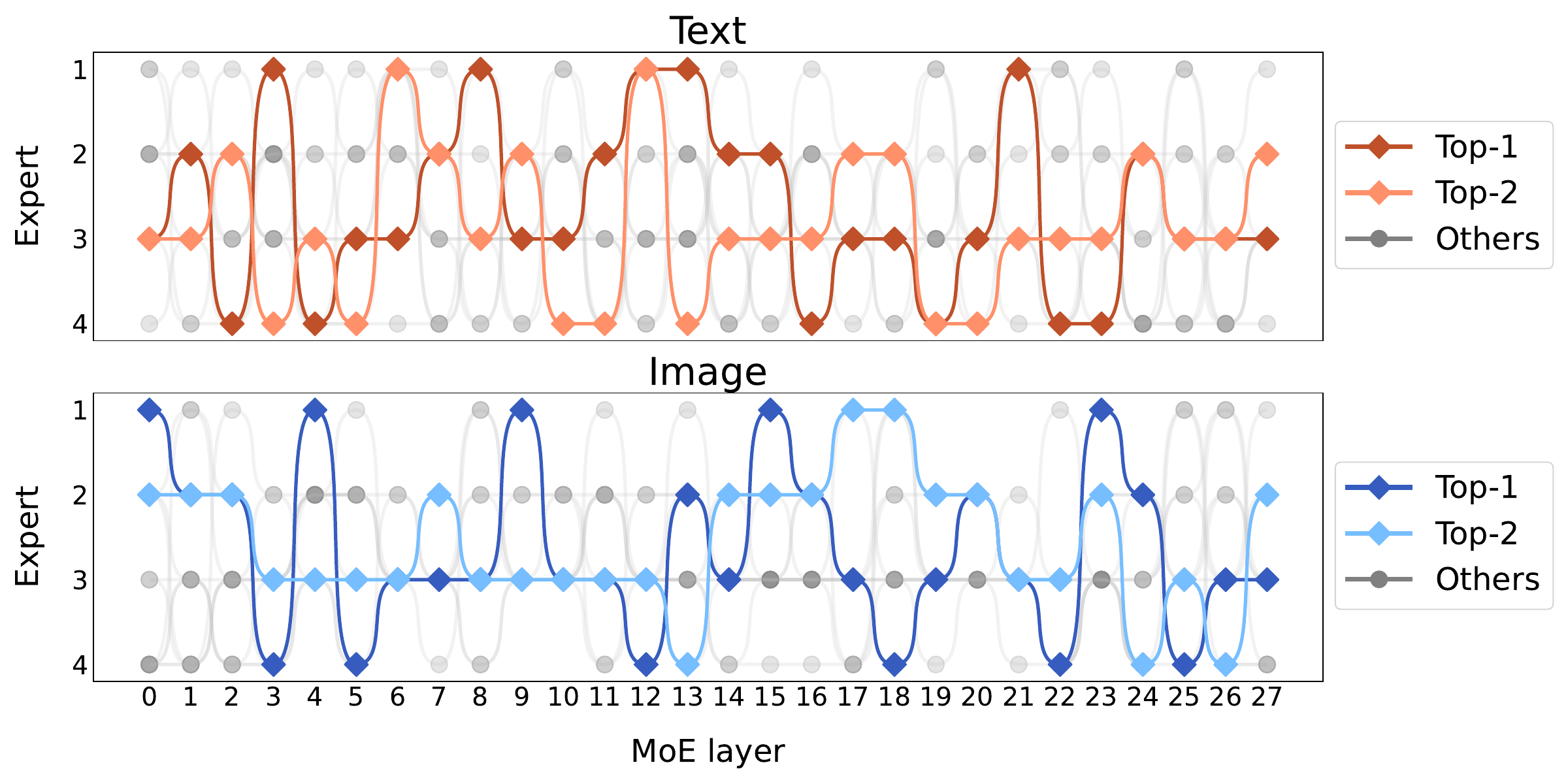}
        \caption{The top-10 pathways for text and image tokens on ChartQA.}
    \end{subfigure}
    
    \begin{subfigure}{0.8\textwidth}
        \centering
        \includegraphics[width=\textwidth]{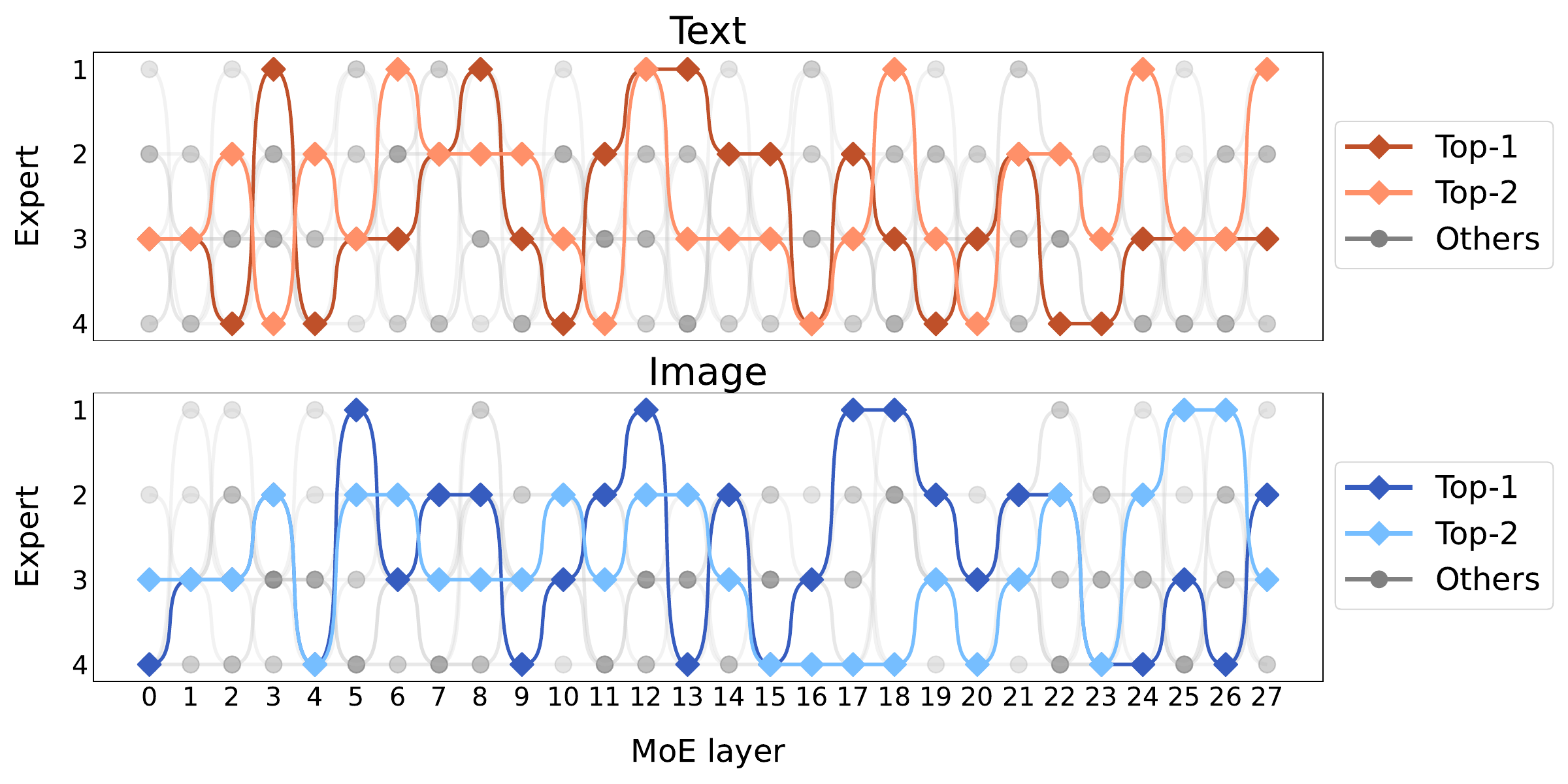}
        \caption{The top-10 pathways for text and image tokens on DocVQA.}
    \end{subfigure}

    \begin{subfigure}{0.8\textwidth}
        \centering
        \includegraphics[width=\textwidth]{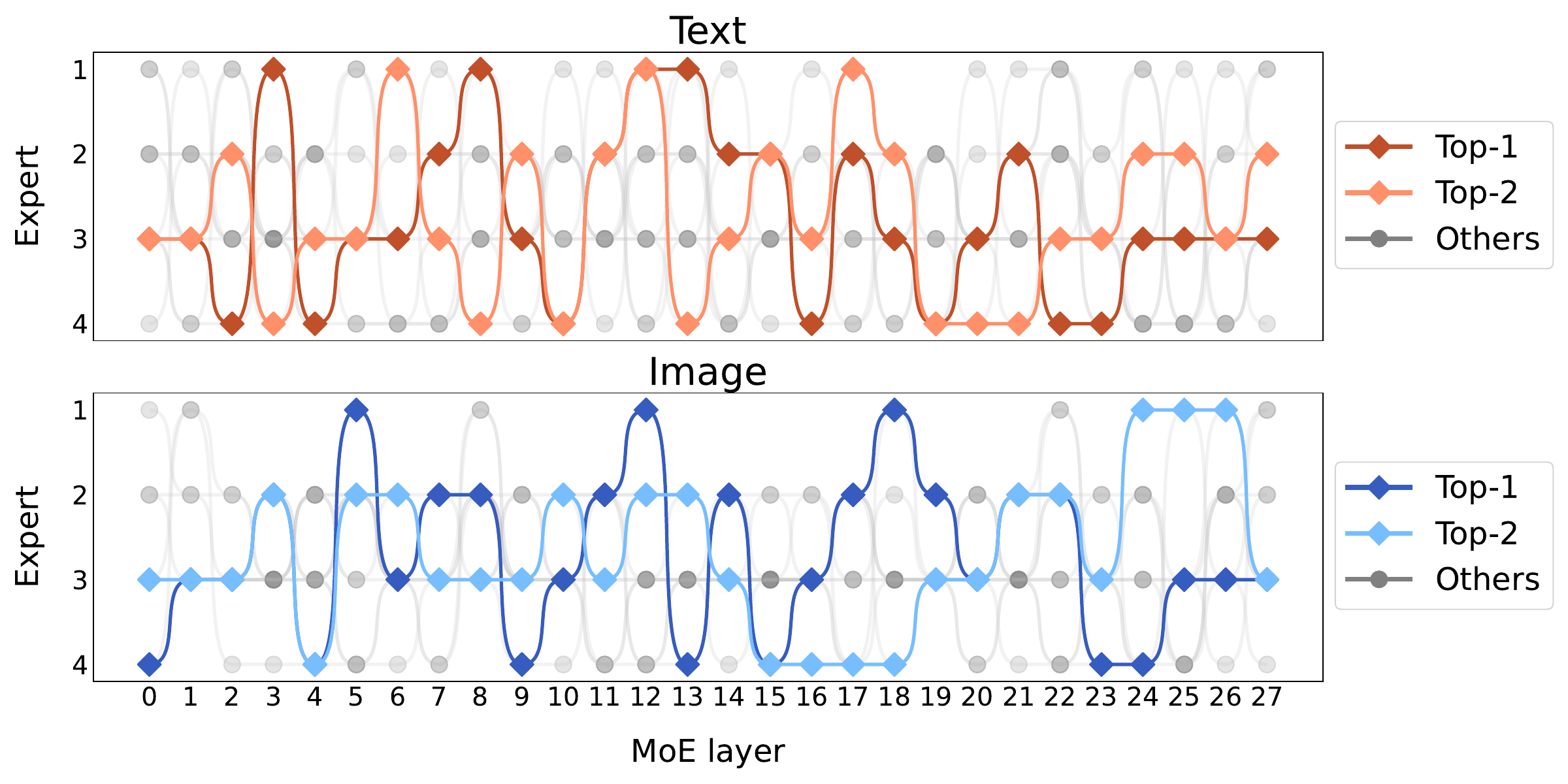}
        \caption{The top-10 pathways for text and image tokens on  InfoVQA.}
    \end{subfigure}
\end{figure*}

\begin{figure*}[h]
    \ContinuedFloat
    \centering
    \begin{subfigure}{0.8\textwidth}
        \centering
        \includegraphics[width=\textwidth]{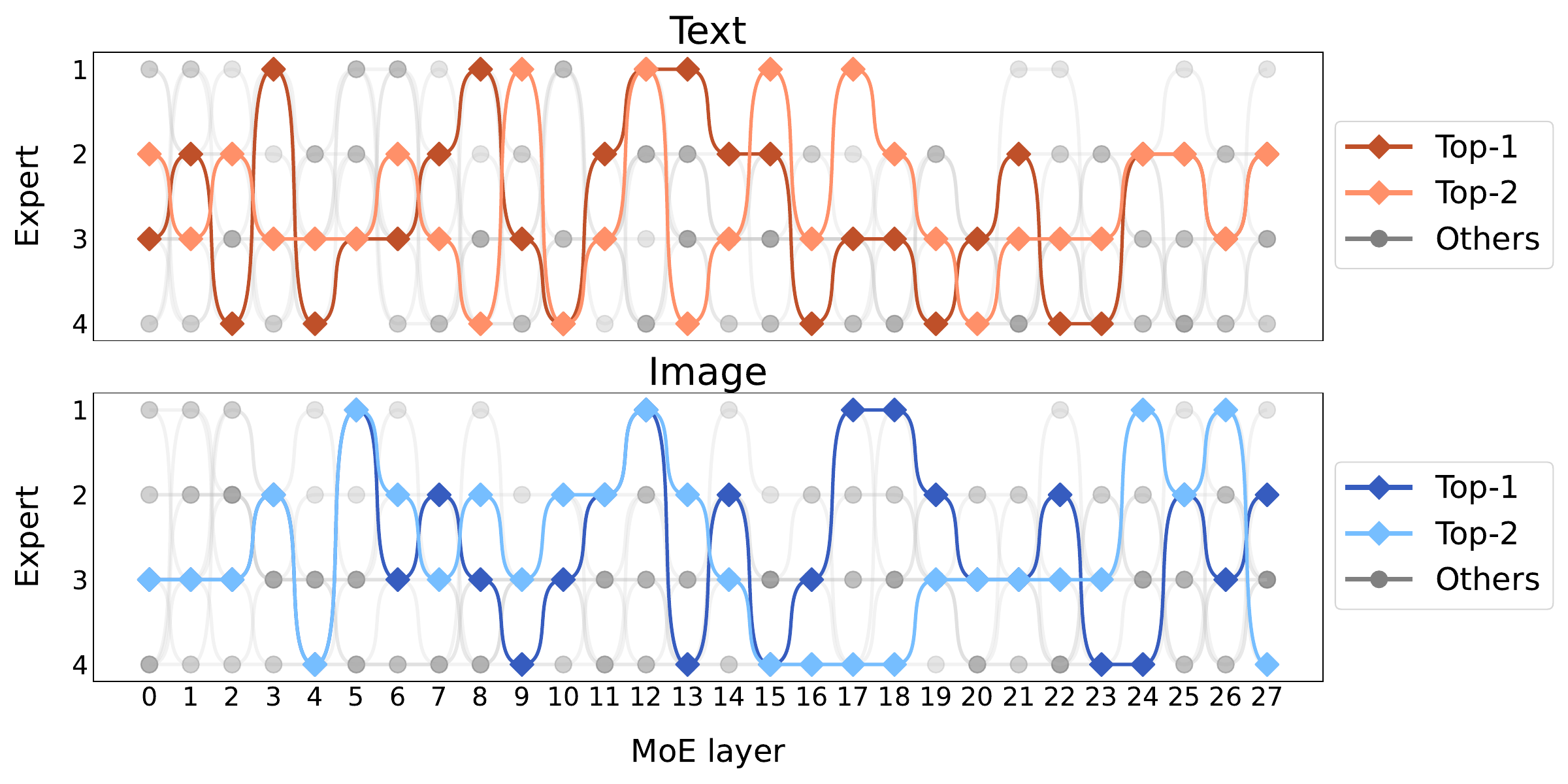}
        \caption{The top-10 pathways for text and image tokens on MMStar.}
    \end{subfigure}
    \caption{Visualization of the top-10 activated pathways for text and image modality on various tasks.}
    \label{ap:vis:fig3}
\end{figure*}

\end{document}